\documentclass[10pt,journal,letterpaper,compsoc]{IEEEtran}
\ifCLASSOPTIONcompsoc
\else
\fi
\ifCLASSINFOpdf
\else
\fi
\usepackage{color}

\definecolor{orange}{rgb}{1,.5,0}
\definecolor{myblue}{rgb}{0,0.5,1}

\usepackage{url}

\usepackage{times}
\usepackage{epsfig}
\usepackage{graphicx}
\usepackage{amsmath}
\usepackage{amssymb}
\usepackage{amsthm}
\usepackage{placeins}

\newcommand{\suppress}[1]{}

\usepackage{multirow}
\usepackage{nicefrac}

\DeclareMathOperator{\atan}{atan}
\DeclareMathOperator{\trace}{trace}

\begin{document}
%
\title{A Region-based Gauss-Newton Approach to \\[2mm]Real-Time Monocular Multiple Object Tracking} 
%
%
%
%

\author{{Henning Tjaden, Ulrich Schwanecke, Elmar Sch\"omer
        and Daniel Cremers}
\IEEEcompsocitemizethanks{\IEEEcompsocthanksitem Henning Tjaden and Ulrich Schwanecke are with the RheinMain University Wiesbaden.  Elmar Sch\"omer is with the Johannes Gutenberg University Mainz. And Daniel Cremers is with the Technical University of Munich.\protect\\
}
\thanks{}
}

\IEEEcompsoctitleabstractindextext{%

\begin{abstract}
We propose an algorithm for real-time 6DOF pose tracking of rigid 3D objects using a monocular RGB camera. The key idea is to derive a region-based cost function using temporally consistent local color histograms. While such region-based cost functions are commonly optimized using first-order gradient descent techniques, we systematically derive a Gauss-Newton optimization scheme which gives rise to drastically faster convergence and highly accurate and robust tracking performance.  We furthermore propose a novel complex dataset dedicated for the task of monocular object pose tracking and make it publicly available to the community. To our knowledge, it is the first to address the common and important scenario in which both the camera as well as the objects are moving simultaneously in cluttered scenes. In numerous experiments - including our own proposed dataset - we demonstrate that the proposed Gauss-Newton approach outperforms existing approaches, in particular in the presence of cluttered backgrounds, heterogeneous objects and partial occlusions. 
\end{abstract}


}

\maketitle

\IEEEdisplaynotcompsoctitleabstractindextext

%
\IEEEpeerreviewmaketitle

\begin{figure*}[!tp] \centering
    \includegraphics[width=0.165\textwidth]{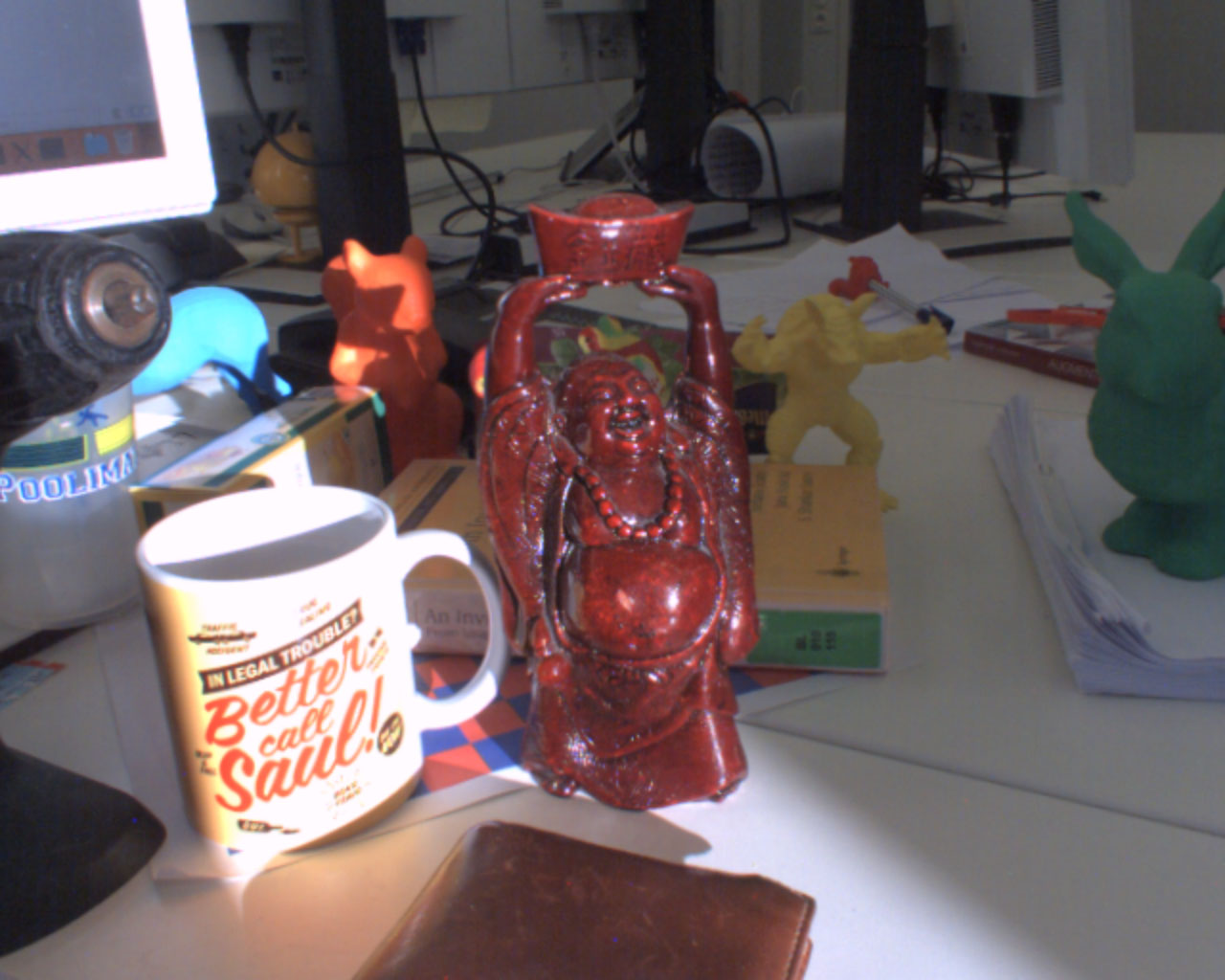}\hfill
    \includegraphics[width=0.165\textwidth]{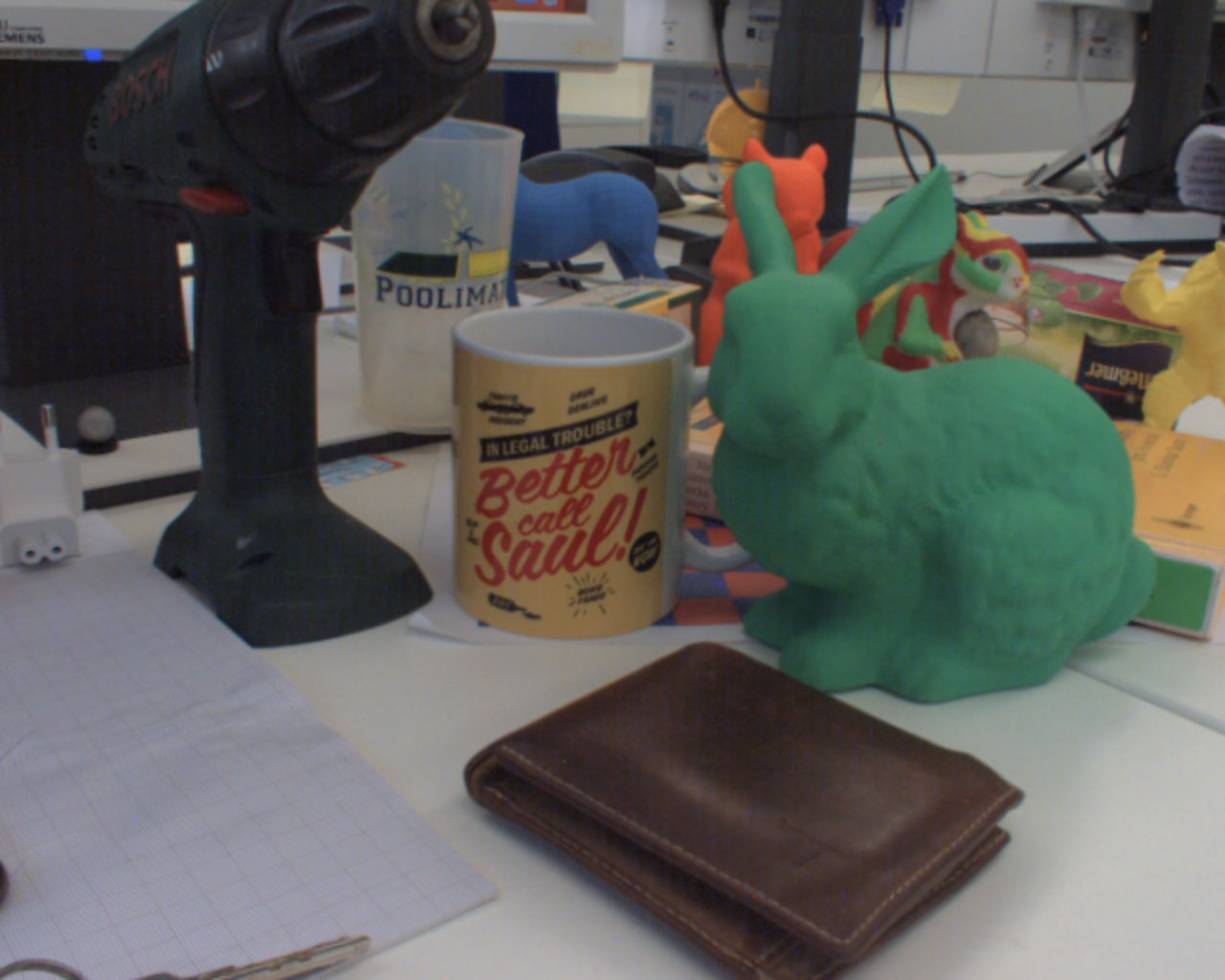}\hfill
    \includegraphics[width=0.165\textwidth]{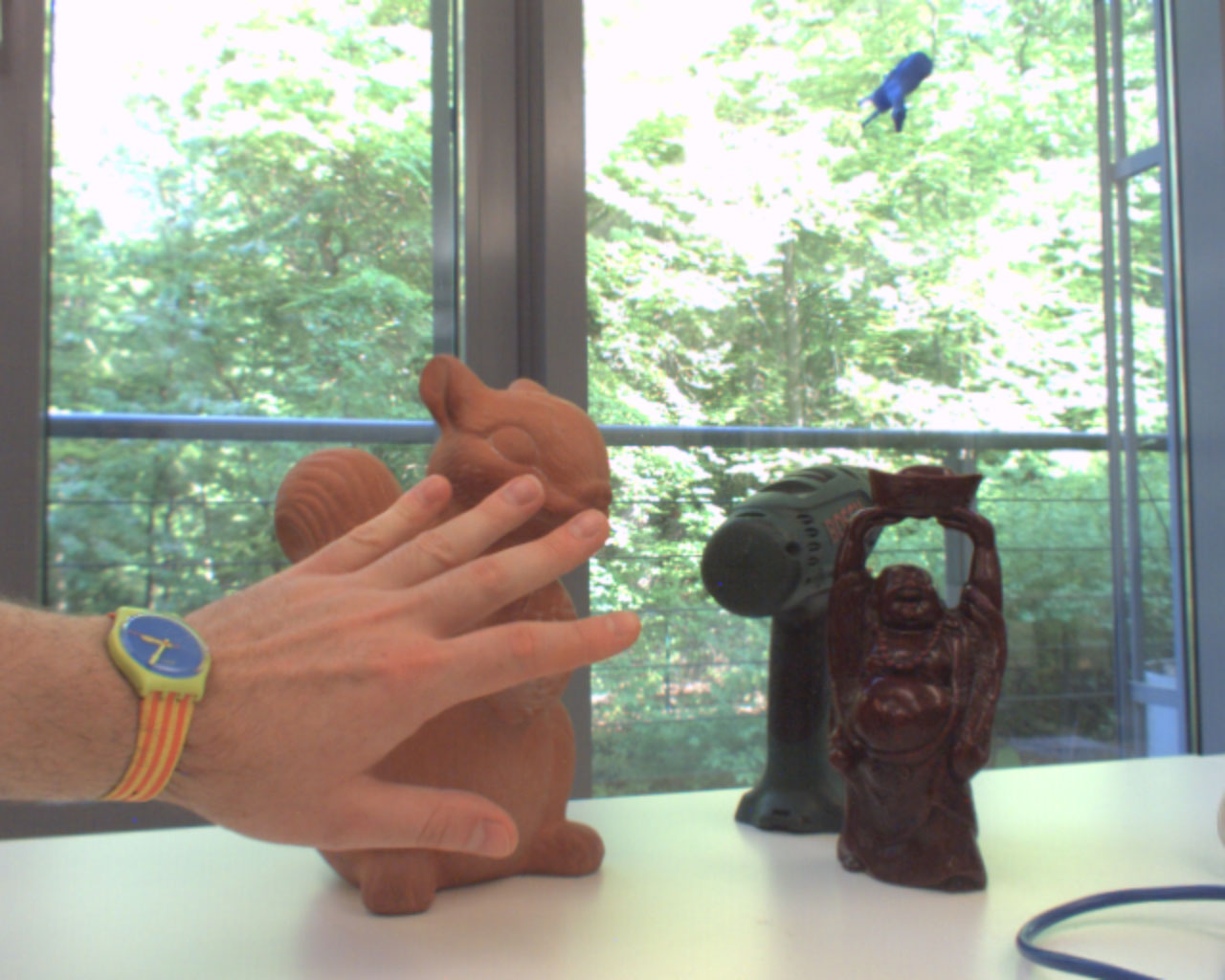}\hfill
    \includegraphics[width=0.165\textwidth]{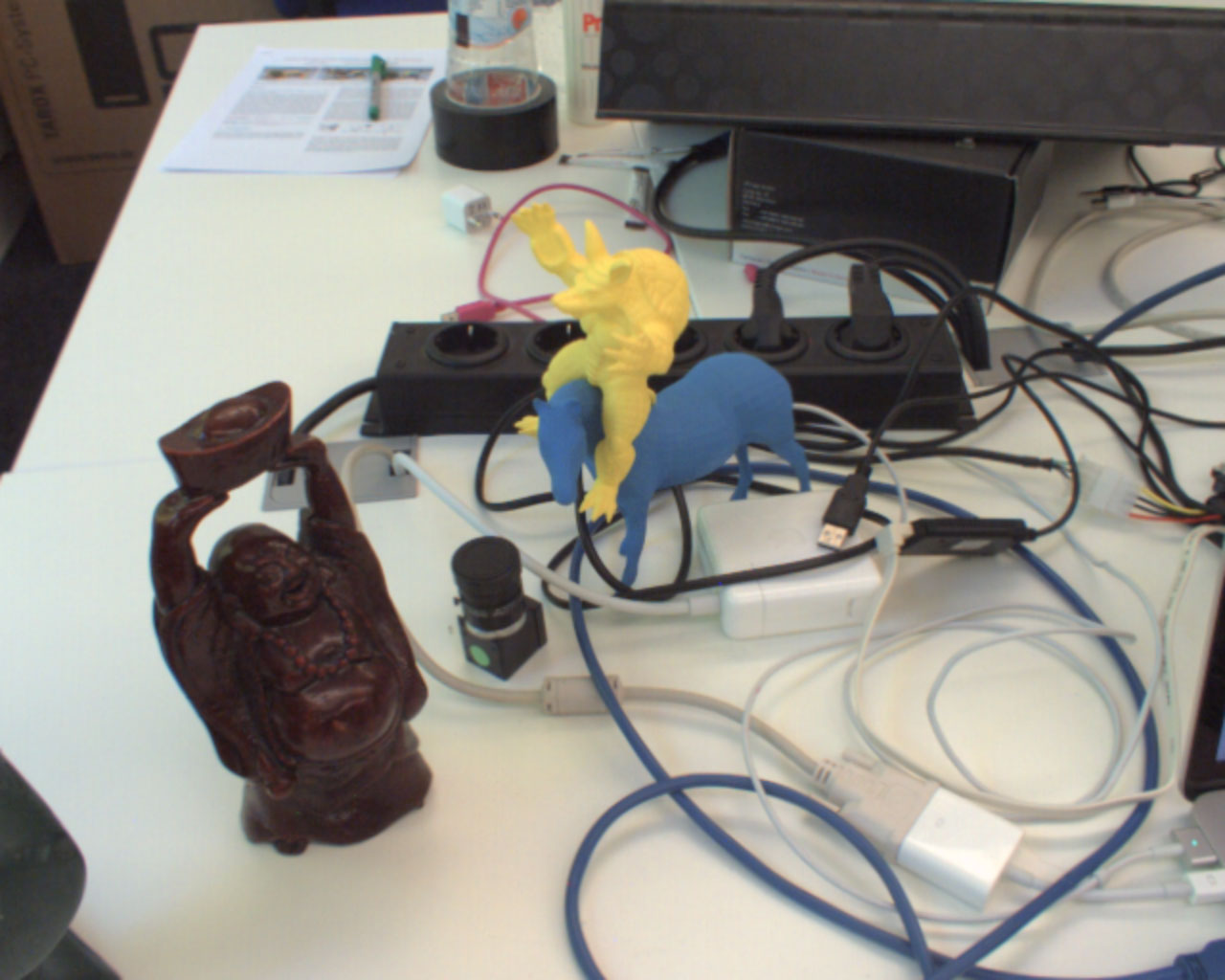}\hfill
    \includegraphics[width=0.165\textwidth]{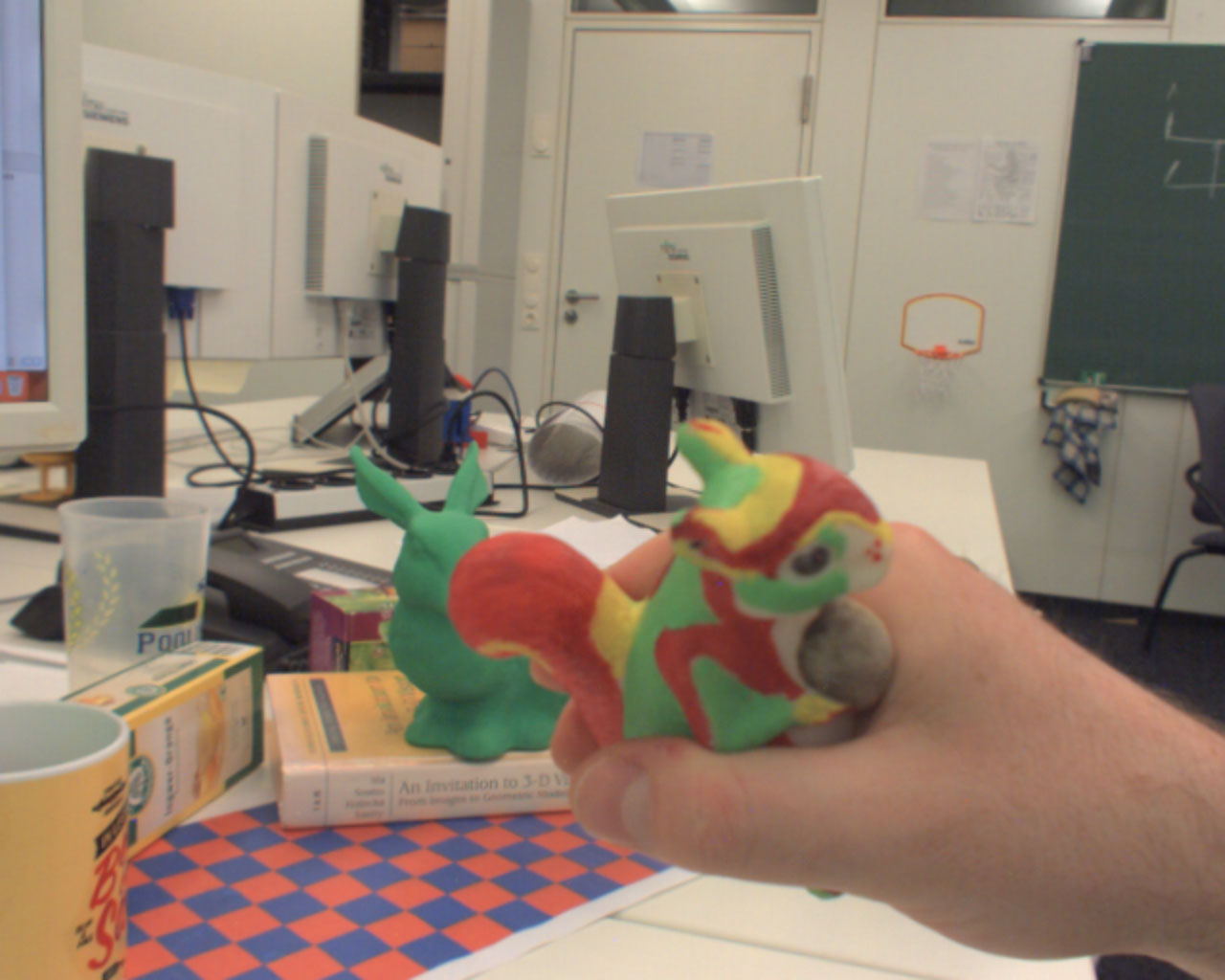}\hfill
    \includegraphics[width=0.165\textwidth]{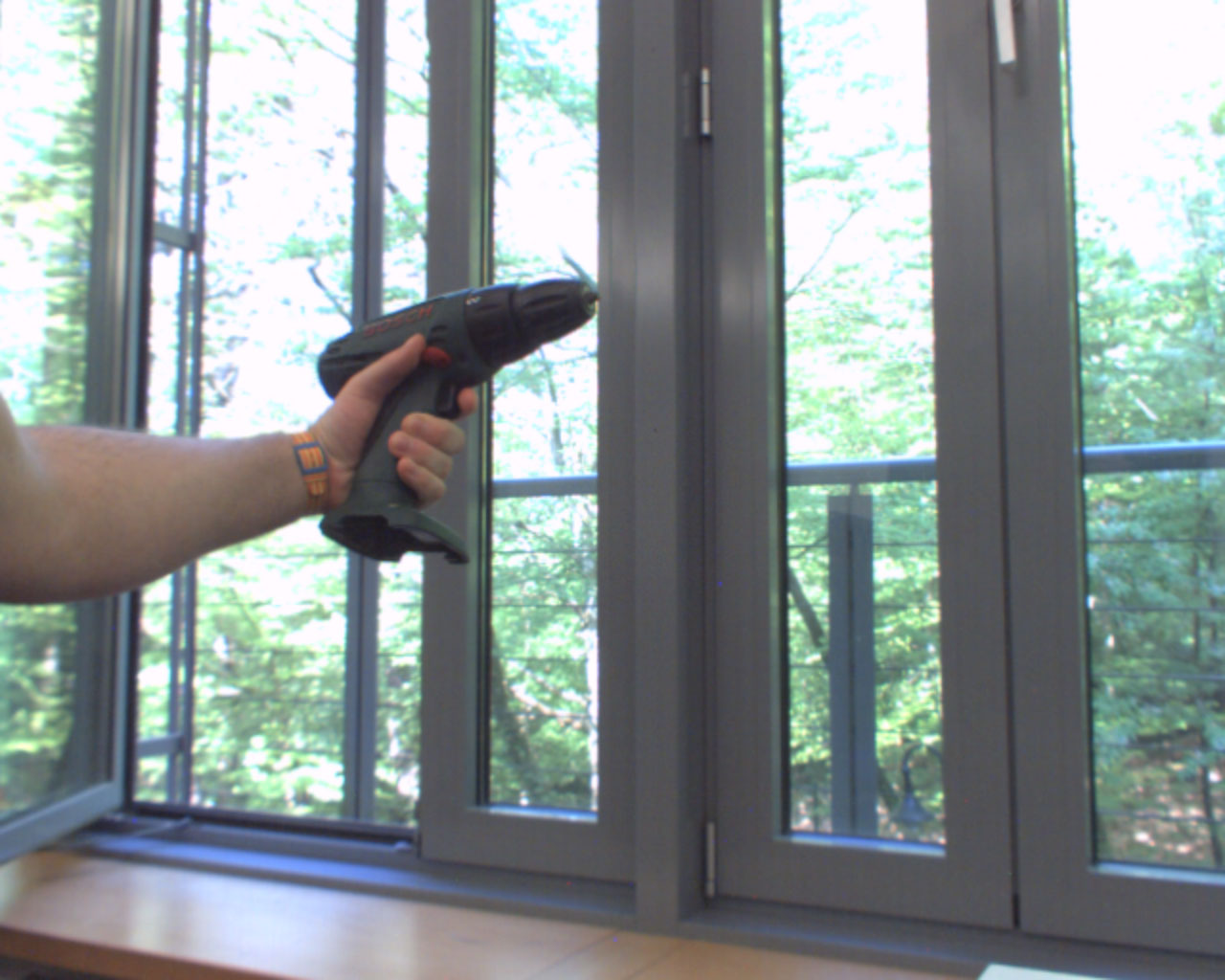}\hfill\\[0.4mm]
    \includegraphics[width=0.165\textwidth]{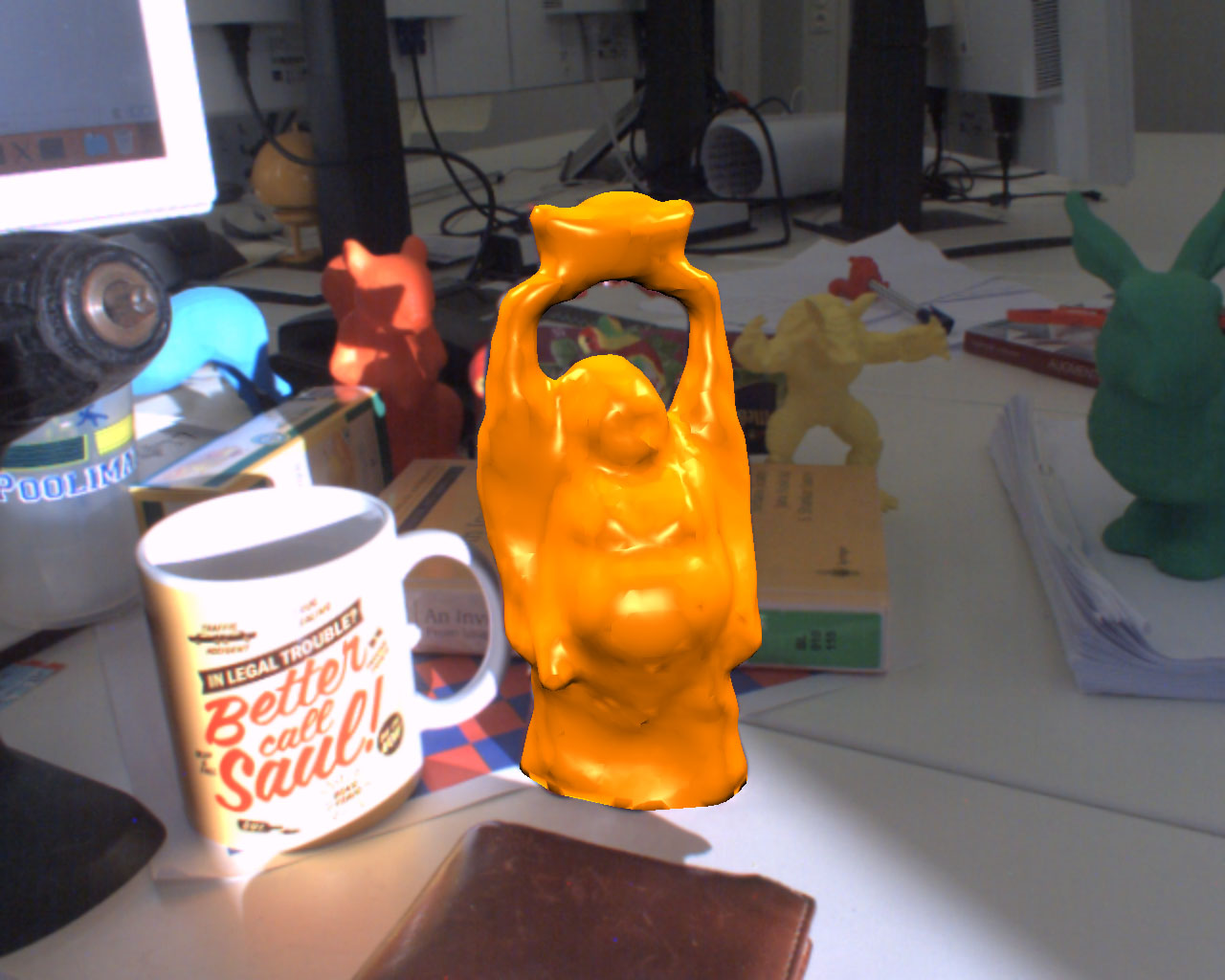}\hfill
    \includegraphics[width=0.165\textwidth]{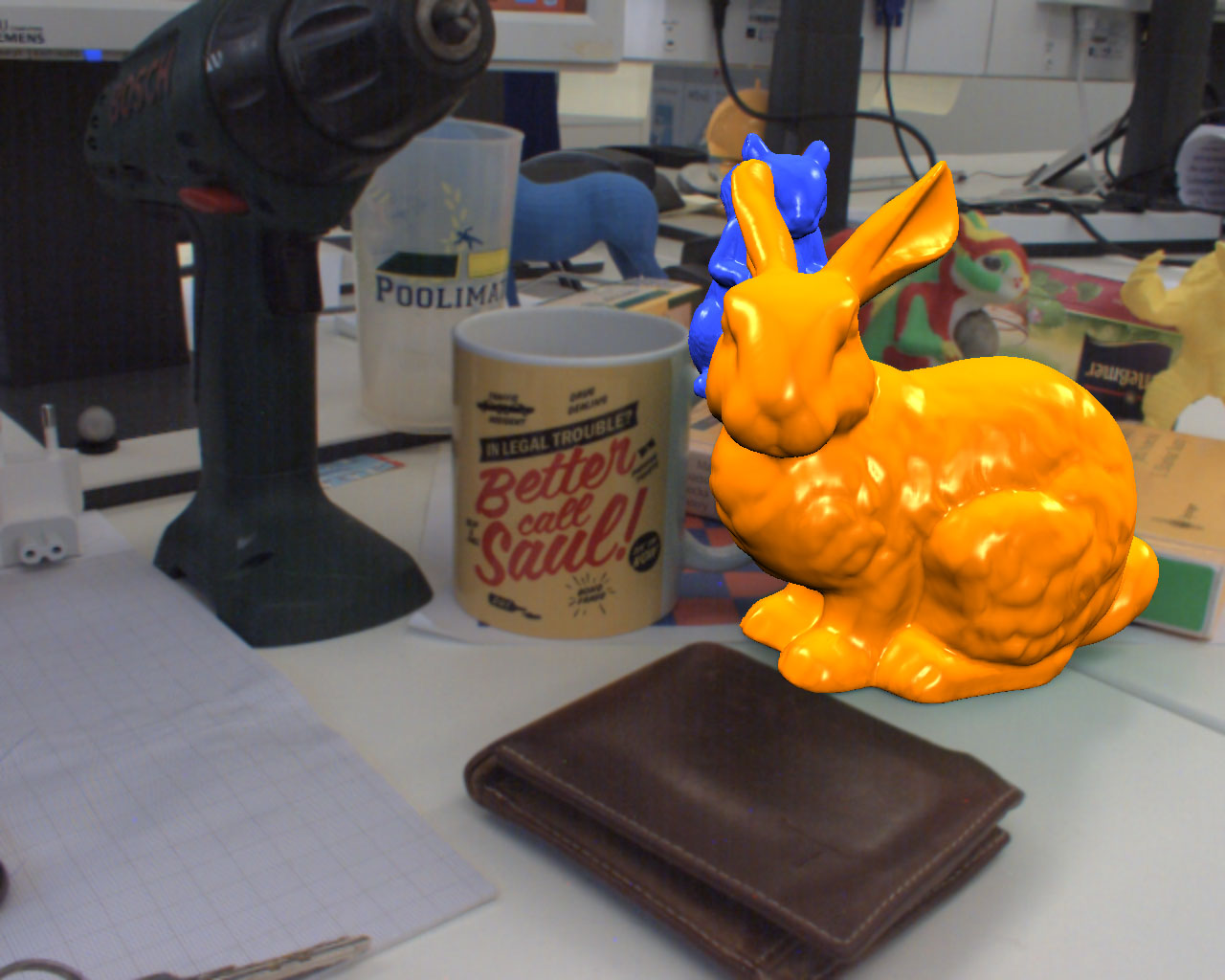}\hfill
    \includegraphics[width=0.165\textwidth]{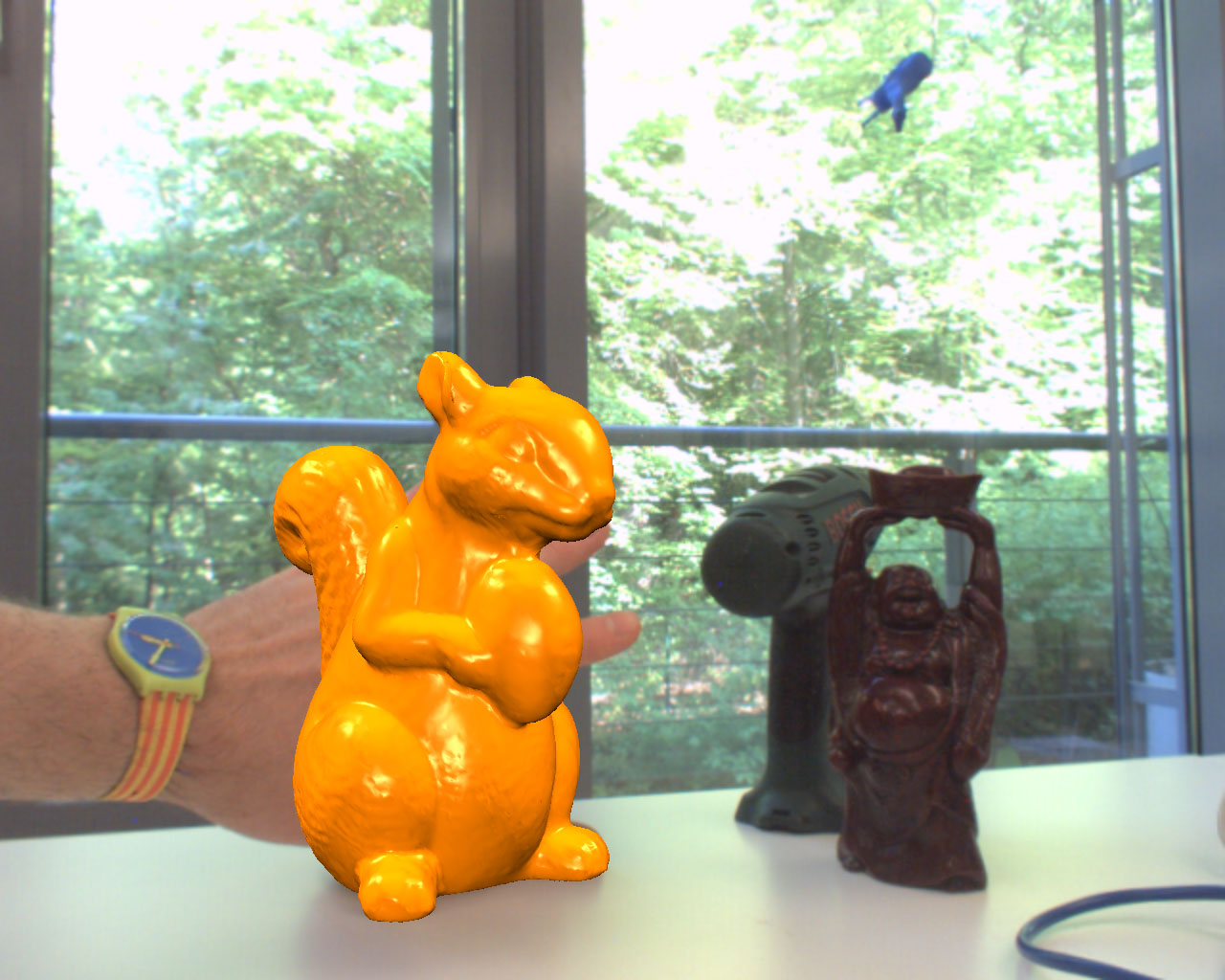}\hfill
    \includegraphics[width=0.165\textwidth]{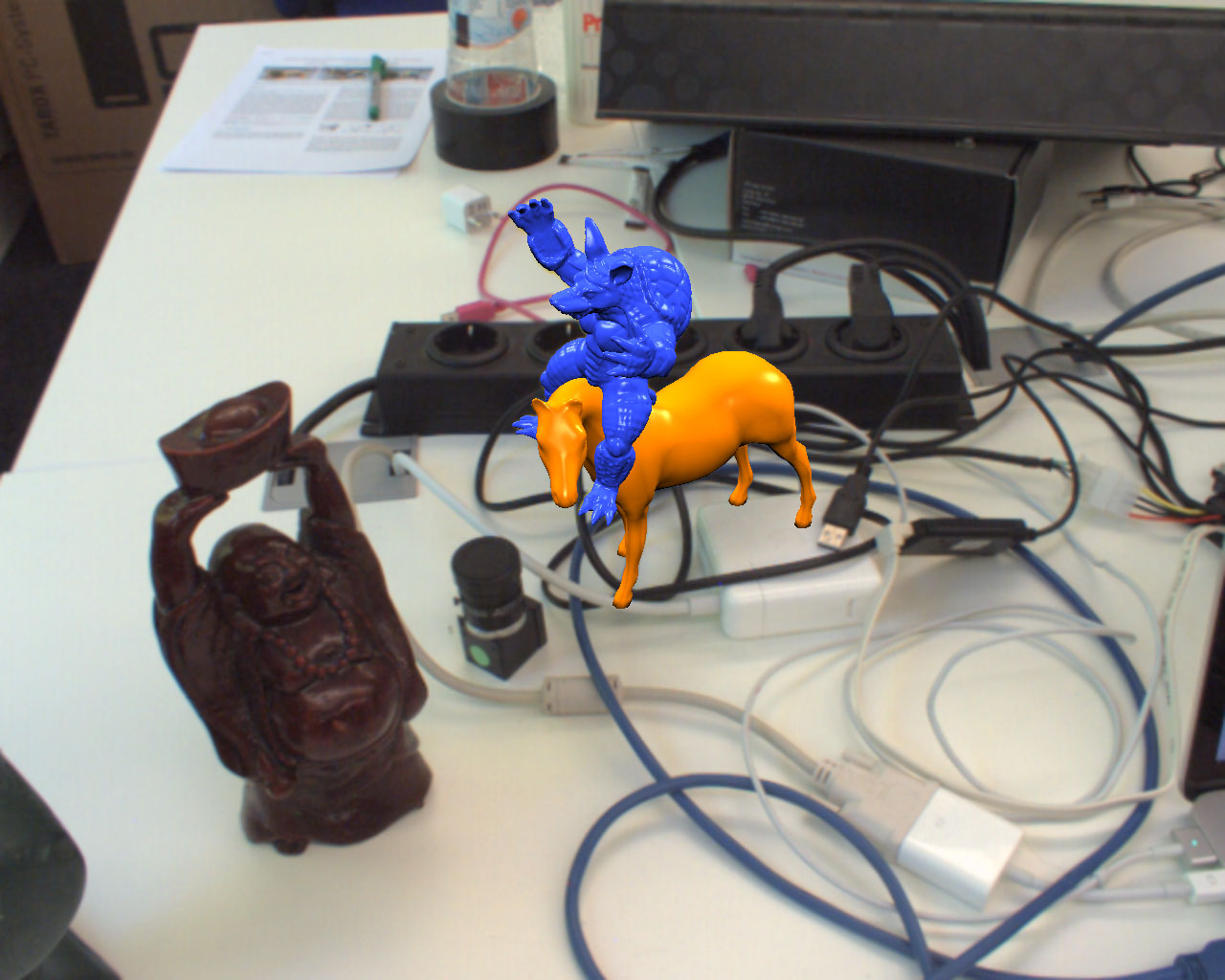}\hfill
    \includegraphics[width=0.165\textwidth]{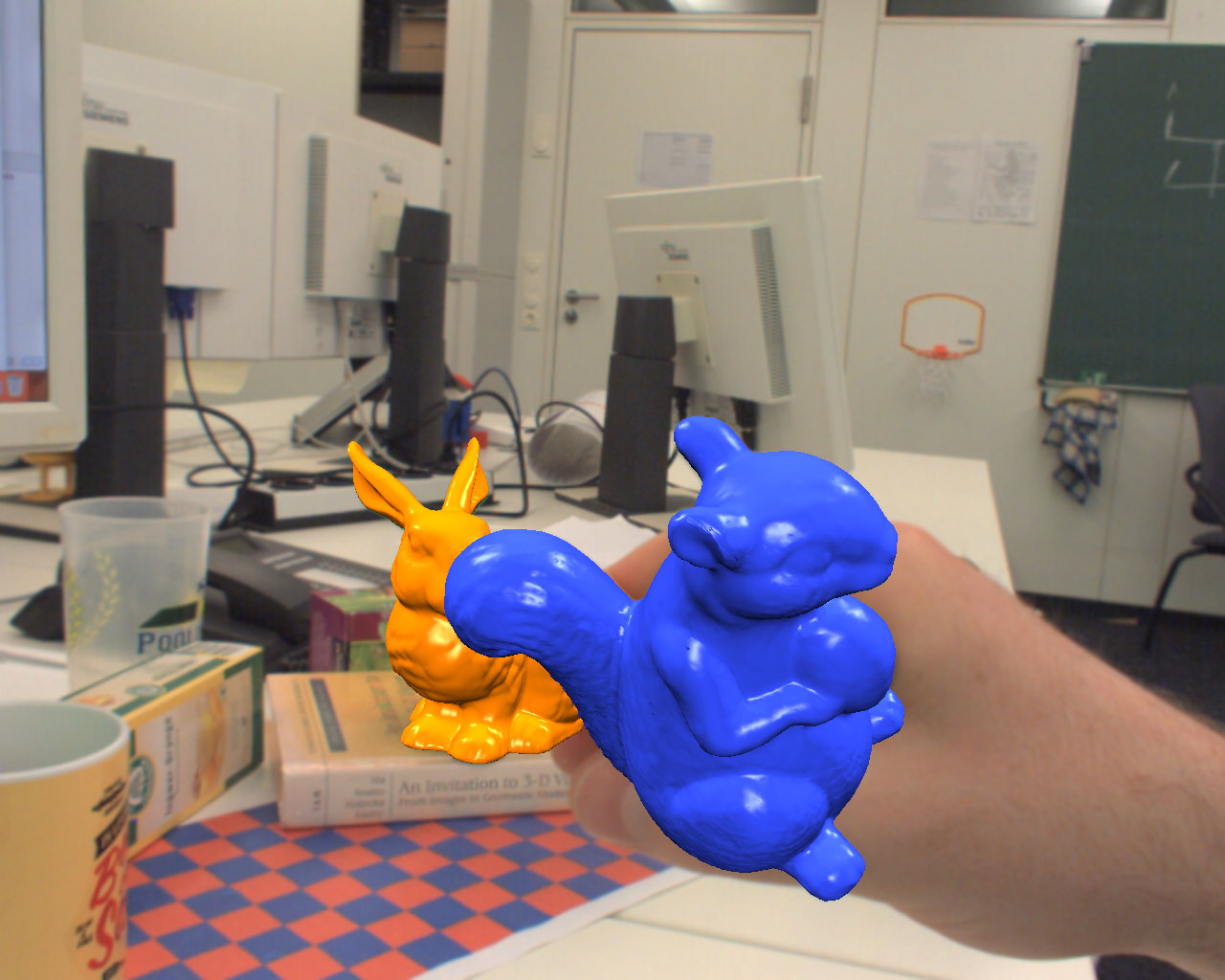}\hfill
    \includegraphics[width=0.165\textwidth]{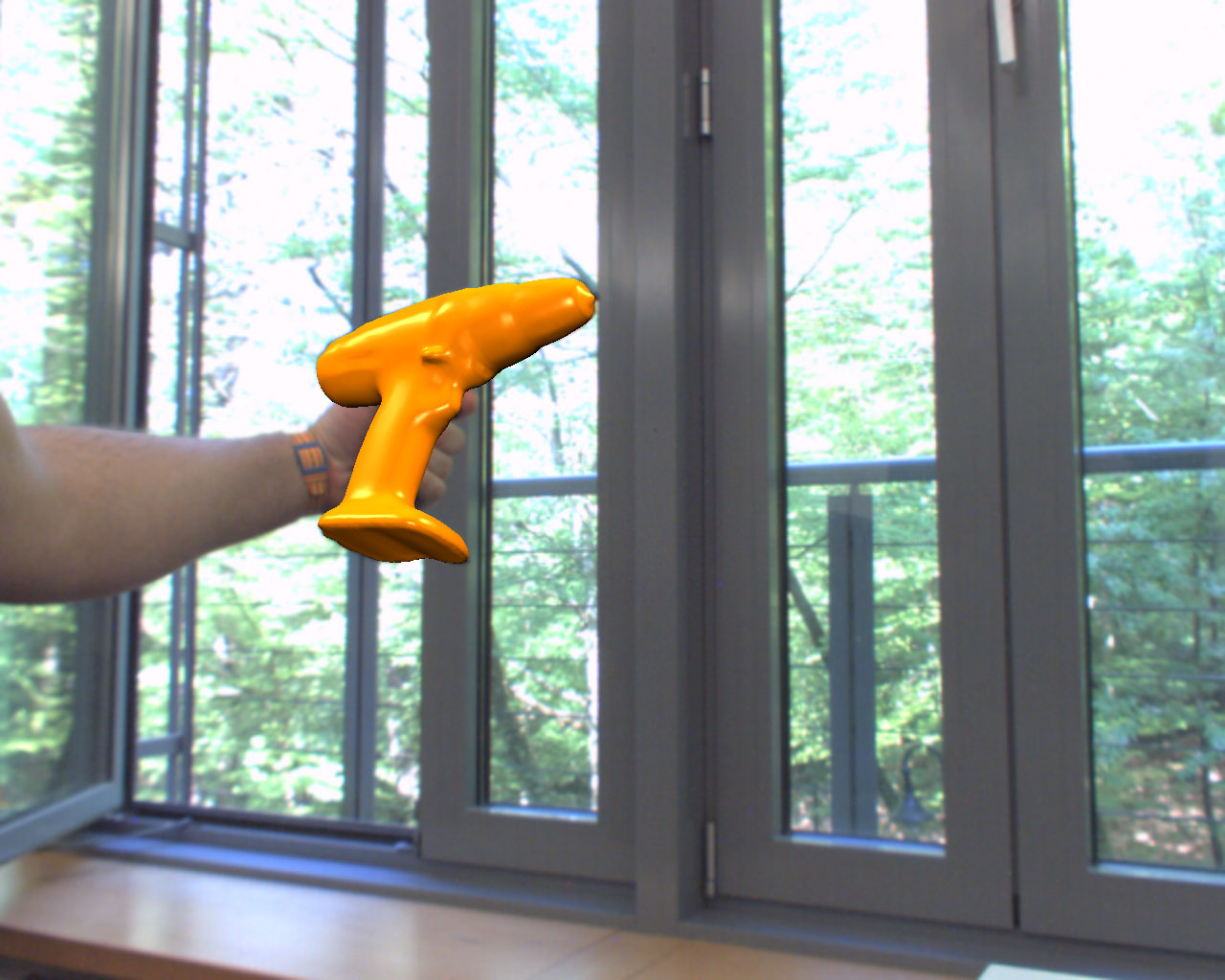}\hfill
    \caption{A few examples of our proposed method estimating the pose of a single or multiple complex objects under different challenging conditions. These include cluttered scenes/backgrounds, strong occlusions as well as direct sunlight. Top: The raw RGB input frames. Bottom: A mixed reality visualization of the tracking result where the input images are virtually augmented with renderings of the corresponding 3D models using the estimated poses. All results were obtained within $\sim$16\,ms per object. This shows the robustness and accuracy of our approach in a variety of situations that are typically difficult for monocular pose estimation.} 
    \label{fig:examples} 
\end{figure*}

\section{Introduction}

Tracking the pose of a rigid object in monocular videos is a fundamental challenge in computer vision with numerous applications in mixed reality, robotics, medical navigation and human-computer interaction~\cite{lepetit05}.  Given an image sequence, the aim is to robustly and accurately determine the translation and rotation of a known rigid object relative to the camera between frames.  While this problem has been intensively studied, accurate real-time 6DOF pose estimation that is robust to background clutter, partial occlusions, motion blur or defocus remains an important open problem (see Figure~\ref{fig:examples}).
 
\subsection{Real-Time Monocular Object Tracking}

Especially in the constantly growing fields of mixed reality and robotics, object pose estimation is typically only one of many complex tasks that must all be computed simultaneously in real-time on often battery powered hardware. Therefore, low runtime and power consumption of dedicated solutions are crucial aspects for them to be practical in such scenarios. In particular the latter can be achieved by using as few sensors as possible e.g. only a single camera. Also as for most other computer vision problems, such monocular approaches are usually the most convenient compared to e.g. multi-camera stereo systems because they keep calibration requirements at a minimum and suffer least from visibility issues.

In the past several different strategies have been proposed to the problem of monocular object pose tracking. One popular approach is to detect intensity gradient-based features such as corners or edges in the images and then perform pose estimation by matching these against a suitable 3D model representation (see e.g.~\cite{vacchetti04, rosten05, hinterstoisser07, wagner10, kim10, lebeda17}). Since here the input image is directly reduced to a sparse set of features a major advantage of such approaches is that they can typically be computed in real-time even on low powered hardware. The main drawback of especially point feature-based methods is, however, that they require the objects to be well-textured in order to show sufficient local intensity variation in the images from all perspectives. In manmade environments this usually significantly limits the variety of suitable objects to those with text or image graphics printed on them. 

For weakly-textured or textureless objects, for many years edge features, describing either their contour or strong intensity gradients of angular structures inside the projected object region, have been shown to be more suitable alternative. Methods relying on image edges are however prone to fail in cluttered scenes that show strong gradients frequently in the background. This potentially causes the pose estimation to end up in local minima (see also e.g.~\cite{seo14}). Feature based methods furthermore struggle with motion blur, low lighting conditions and increasing distance to the camera, generally causing the features to appear less distinct in the images.

More recently, so-called region-based methods have gained increasing popularity (see e.g.~\cite{rosenhahn06, dambreville08, brox10, schmaltz11, prisacariu12, hexner16, tjaden17}) since they are potentially suitable for a large variety of objects in complex scenarios regardless of local intensity gradients, i.e. their texture. These approaches assume differing image statistics between the object and the background region. Based on a suitable statistical appearance model as well as a 3D shape prior (i.e. a 3D model of the object), here pose estimation essentially works by aligning two silhouettes. Here, the target is the extracted object's silhouette in the current image using the segmentation model while the other is rendered synthetically from the shape prior parametrized by the sought pose. The discrepancy between these two shapes is then minimized by changing the pose parameters used for the synthetic projection. In return, given the objects pose in the current frame, the rendered silhouette provides an accurate pixel-wise segmentation mask that is typically used for updating the foreground and background statistics of the appearance model in order to dynamically adapt to scene changes. Therefore, given the pose in the first frame of an image sequence  allows to initialize the statistical model. Pose tracking is then performed recursively in an interleaved manner by first estimating the pose based on that in the previous frame and then updating the segmentation model afterwards using the mask information in the current frame.

Very recently the first object pose tracking approach that utilizes deep learning techniques to robustly handle occlusions and lighting changes has been proposed in~\cite{crivellaro18}. However, it requires manual labeling of so-called \textit{stable parts} for each object individually. Also, this approach, even when using a powerful GPU, is not real-time capable.

For the sake of completeness we also want to mention that ever since so-called RGB-D cameras (also known as \textit{depth sensors}) have become available as consumer hardware a couple of years ago, another major category of pose tracking algorithms has emerged that rely on these devices (see e.g.~\cite{choi13, krull14, kehl17a, ren17, tan17}). These sensors potentially measure the per pixel distance to the camera in real-time by combining an infrared light emitter, that actively projects light onto the scene, with an monochrome camera and often an RGB camera in a rigid stereo setup. Due to the additional depth modality such methods commonly outperform those only based on monocular RGB image data. However, due to the active lighting strategy they only operate properly within about 10 meters proximity to the device, generally struggle in the presence of sunlight and shiny surfaces and have a higher power consumption than a regular camera. We therefore do not include methods requiring such sensors as related works since we do not consider them sufficiently comparable to the monocular setting.

\subsection{Related Work}
 
When only using a single regular RGB camera, to our best knowledge region-based approaches relying on statistical level-set segmentation~\cite{cremers07} are currently achieving state-of-the-art performance for the task of 6DOF object pose tracking. For this reason and due to the large amount of literature on other methods in this domain, here we strictly focus on work that is directly related to our proposed region-based approach. By also presenting a newly constructed complex dataset, we address a gap in literature regarding current publicly available datasets for monocular 6DOF pose tracking of 3D objects. Thus, the seconds part of this section gives a comprehensive overview of related datasets and their shortcomings compared to the one we present in this work. 

\textbf{Region-based Pose Tracking Methods} -- Early region-based pose tracking methods were not real-time capable~\cite{rosenhahn06, brox10, schmaltz11} but already showed the vast potential of the general strategy, by presenting promisingly robust results in many complex scenarios. In these works image segmentation was based on level-sets together with pixel-wise likelihoods used to explicitly extract the object's contour in each camera frame. Here, pose estimation was based on an iterative closest points (ICP) approach by solving a linear system, set up from 2D-to-3D point correspondences between the extracted contour and the 3D model. These correspondences are re-established after each iteration in the 2D image plane to the evolving synthetic contour projection.

In~\cite{prisacariu12} the authors presented PWP3D, the first region-based approach that achieved real-time frame rates (20--25\,Hz) by relying heavily on GPGPU acceleration. Here, pose estimation is performed using a pixel-wise gradient-based optimization scheme similar to the variational approach suggested in~\cite{dambreville08}. But instead of separately integrating over the foreground and background region, PWP3D uses a level-set pose embedding in a cost function similar to the very early methods above in order to simplify computations and make it real-time capable. Additionally, based on the idea presented in~\cite{bibby08} the previously proposed pixel-wise likelihoods were exchanged for pixel-wise posterior probabilities which have been shown to provide a wider basin of convergence.

There have been several successive works recently that build upon the general concept of PWP3D. These mostly address two main potential improvements of the original algorithm. The first being the first-order gradient descent used for pose optimization involving four different fixed step sizes that have to be adjusted experimentally for each model individually. It also uses a fixed number of iterations to maintain real-time performance and thus suffers from commonly related convergence issues. This optimization was replaced in~\cite{tjaden16} with a second-order so-called Gauss-Newton-like optimization where the Hessian matrix is approximated from first-order derivatives based on linearized twist parametrization. This strategy vastly enhanced the convergence properties resulting in significantly increased robustness towards fast rotations and scale changes. Furthermore, by performing this optimization in an hierarchical coarse-to-fine manner the overall runtime of the proposed mainly CPU-based implementation (it only uses OpenGL for rendering) was reduced to achieve frame rates of 50--100\,Hz for a single object on a commodity laptop. However, in~\cite{tjaden16} the optimization strategy was discovered from empirical studies and thus not properly derived analytically.

Another CPU-based approach was presented in~\cite{prisacariu15} that achieves around 30\,Hz on a mobile phone by using an hierarchical Levenberg-Marquardt optimization strategy for the translation parameters and approximating the level-set related computations. However the main speed-up was enabled by including the phone's gyroscope to obtain the rotation estimate that is only corrected for drift every tenth frame by a single gradient descent step. Due to this sensor fusion the method presented in~\cite{prisacariu15} can technically not be considered a monocular solution and is furthermore restricted to application scenarios in which the phone moves around a static object. 

The second main disadvantage of the original PWP3D method is the rather simple segmentation model based on global foreground and background color histograms which is prone to fail in cluttered scenes. Therefore the authors of~\cite{zhao14} introduce a boundary constraint for improvement, that is however not real-time capable. In~\cite{hexner16} based on the idea presented in~\cite{lankton08}, a localized segmentation model was proposed that also relies on pixel-wise posteriors but uses multiple local color histograms to better capture spatial variations of the objects. However, in~\cite{hexner16} this approach was neither evaluated for pose tracking in video sequences nor shown to be real-time capable. 

In the latest work on region-based pose estimation~\cite{tjaden17} presents a real-time capable implementation of~\cite{hexner16} in combination with the ideas of~\cite{tjaden16} and further extends the segmentation model by introducing temporally consistency and a pose detection strategy to recover from tracking losses. The resulting algorithm currently achieves state-of-the-art real-time tracking performance by using the Gauss-Newton-like optimization and a segmentation model based on so-called temporally consistent local color histograms. 

The Gauss-Newton-like optimization was recently also adopted in~\cite{kehl17a} which directly builds up on~\cite{tjaden16}. Here, an extended cost function with respect to the depth modality of an RGB-D device is derived in order to improve on both the robustness of the pose estimation as well as the object segmentation in cluttered environments. To obtain the object pose it is suggested to combine the Gauss-Newton-like approach for the RGB-based term with a standard Gauss-Newton strategy for the depth-based term in a joint optimization. 

\begin{figure}[!tp] \centering
    \includegraphics[width=0.497\columnwidth]{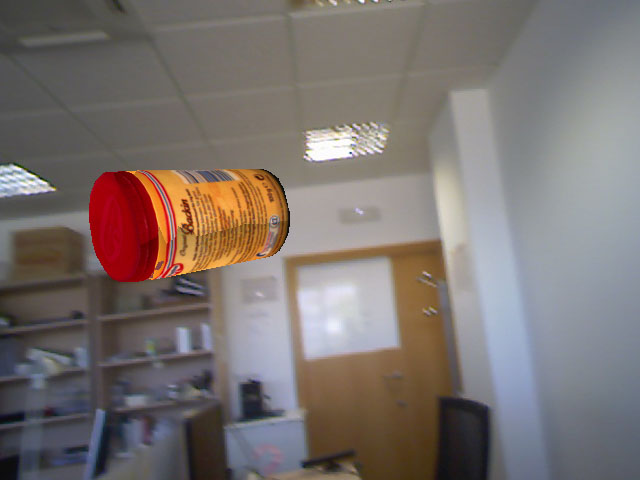}\hfill
    \includegraphics[width=0.497\columnwidth]{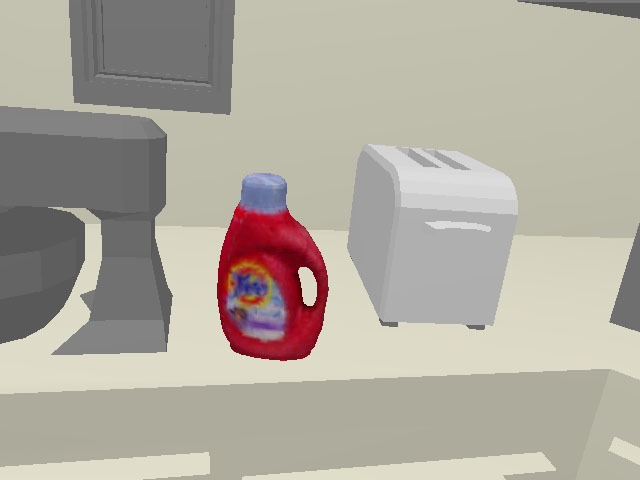}\hfill\\[0.4mm]
    \includegraphics[width=0.497\columnwidth]{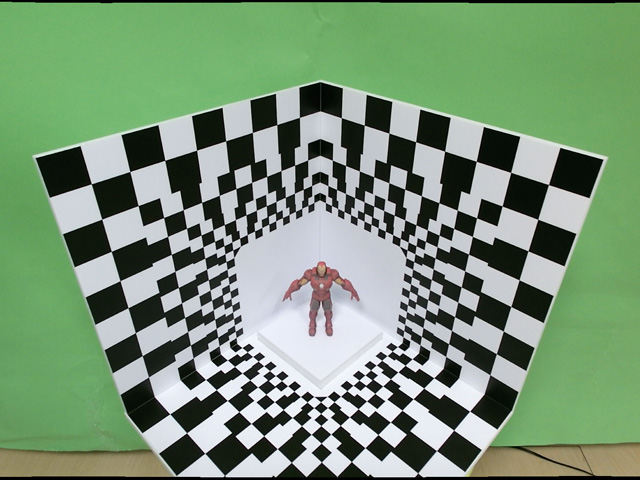}\hfill
    \includegraphics[width=0.497\columnwidth]{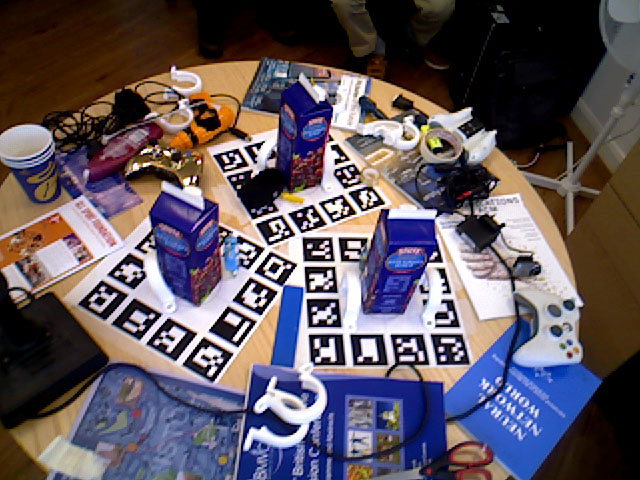}\hfill
    \caption{Example images extracted from different related pose tracking datasets. Top left:~\cite{pauwels13}, top right:~\cite{choi13}, bottom left:~\cite{wu17} (cropped to fit aspect ratio) and bottom right:~\cite{tejani14}.} 
    \label{fig:datasets} 
\end{figure}

\textbf{Object Pose Tracking Datasets} -- There are several different aspects that can potentially be covered by an object pose tracking dataset. One important feature is the type and complexity of motion included in the respective image sequences. Here, the \textit{type} of motion refers to whether they include either movement of only the camera, only the object or both simultaneously. This is particularly important because it has a direct impact on e.g. the intensity gradients (i.e. shading) inside the object region that depend on the lighting in the scene and how quickly the background changes. Another aspect is whether the light sources in the scene are moving. It is closely related to the case of object motion but typically has an even stronger impact on the objects appearance (e.g. self-shadowing). With regard to the previously discussed gradient-based image features, it should also be considered to include both well-textured as well as weakly textured objects. Datasets can furthermore contain a single or multiple objects, occlusions, different amount of background clutter and motion blur to simulate common problems in real scenarios. 

Another essential question when generating an object tracking dataset is how the ground truth pose information is obtained for each frame, since this is basically a chicken-egg problem. One popular and straightforward approach is to render synthetic image sequences from artificial and thus fully controllable 3D scenes. However, the resulting images often lack of photo-realism and require sophisticated rendering techniques and expert knowledge in 3D animation. On the other hand, in case of real-data image sequences typically some sort of fiducial markers are placed in the scene to provide a reference pose from all perspectives independent of the rest of the scene. This requires the relative pose between markers and object to remain static at all times. Datasets of this kind thus often only contain motion of the camera unless markers are also attached to the object which, however, changes its appearance in an unnatural and undesired way. 

In the context of 6DOF object pose tracking a third, sort of intermediate strategy can be applied, where semi-synthetic image sequences are created, combining advantages of both previous solutions. Here, animated renderings of a realistically textured 3D model are composed with a real image sequence providing for a background in each frame (see Section~\ref{sec:ourdataset} for details). This technique has been used for the \textit{Rigid Pose} dataset presented in~\cite{pauwels13}, which we consider most closely related to the one we propose. Here semi-synthetic stereo image pair sequences of six different objects are provided, each available in a noise-free, a noisy and an occluded version. However, five out of the six objects used within this dataset are particularly well textured. Also the objects are rendered using a Lambertian illumination model without including any directional light source, meaning that the intensity of corresponding pixels between frames does not change. These renderings are furthermore simply pasted onto real images without e.g. blurring their contours in order to smooth the transition between the object and the background (e.g. Figure~\ref{fig:datasets}, top left).  

Then there is the \textit{RGB-D Object Pose Tracking} dataset of~\cite{choi13} that contains two real-data and four fully synthetic sequences including four different objects that are static in the scene. However, ground truth information is only provided for the four synthetic sequences which were primarily designed for depth-based tracking. Here, apart from the respective object itself the rest of the scene is very simple and completely texture-less, which is why the resulting RGB color images look very artificial overall (e.g. Figure~\ref{fig:datasets}, top right).

Very recently the \textit{OPT} dataset was presented in~\cite{wu17}, being the most complex 6DOF object pose tracking dataset yet. It contains multiple real-data RGB-D sequences of six 2D patterns and six 3D objects that vary in texture and complexity. The images were captured with a camera mounted on a robot arm that moves around each single object at different speeds and varying lighting conditions. The dataset thereby even covers scenarios with a moving light source and contains motion blur. Despite its complexity the data does not include object motion, background clutter or occlusions, since in all sequences the object is placed statically in front of an entirely white background surrounded by a passive black and white marker pattern (e.g. Figure~\ref{fig:datasets}, bottom left). 

Lastly, there is the dataset of~\cite{tejani14}, that contains six real-data RBG-D sequences involving six different objects and partial occlusions. In each sequence multiple static instances of the same object are placed on a cluttered table each surrounded by a marker pattern (e.g. Figure~\ref{fig:datasets}, bottom right). Therefore, the dataset also only contains movement of the camera. Although this dataset does provide test sequences of consecutive video frames, it was primarily designed for the task of 6DOF object pose detection. In that case the object pose is supposed to be recovered from only a single image as opposed to an image sequence in case of pose tracking. Other pose detection datasets (e.g.~\cite{hinterstoisser12a}) typically do not contain any consecutive frames at all and can therefore not be used for pose tracking, although the two tasks are actually strongly related. 

To summarize, there are currently only a few datasets that have been created explicitly for the task of monocular 6DOF pose tracking. Of those available,~\cite{pauwels13, choi13, tejani14} are relatively small and do not cover many of the initially mentioned aspects. The most complex data set currently available~\cite{wu17} unfortunately also does not simulate scenarios in which both the object and the camera are moving (e.g. a hand-held object in a cluttered environment), which we are targeting here.




\subsection{Contribution}

In this work, we derive a cost function for estimating the 6DOF pose of a familiar 3D object observed through a monocular RGB-camera. Our region-based approach involves a statistical image segmentation model built from multiple overlapping local image regions along the contour of the object in each frame. The core of this model are temporally consistent local color histograms (\textit{tclc-histograms}) computed from these image regions, each anchored to a unique location on the object's surface, which allows to update them with each new frame.

While traditionally such cost functions have been optimized by means of gradient descent, we derive a suitable Gauss-Newton optimization scheme. This is not straightforward since the cost function is not in the traditional nonlinear least-squares form. In numerous experiments, we demonstrate the effectiveness of our approach in different complex scenarios, including motion of both the camera and the objects, partial occlusions, strong lighting changes and cluttered backgrounds in comparison to the previous state of the art. 

This work builds up on two prior conference publications~\cite{tjaden16,tjaden17}. It expands these works on several levels:  Firstly, we propose a systematic derivation of a Gauss-Newton optimization by means of reformulating the optimization problem as a reweighted nonlinear least-squares problem. This further improves the convergence rate and thus the tracking robustness significantly compared to the previous Gauss-newton-like scheme of~\cite{tjaden16}. Secondly, we explain how our method using tclc-histograms can be extended to multi-object tracking and thereby handle strong mutual occlusions in cluttered scenes and demonstrate its potential for real-time applications of mixed-reality scenarios. Thirdly, we propose a novel large semi-synthetic 6DOF object pose tracking dataset, that covers most of the previously mentioned important aspects. In our opinion this closes a gap in the current literature on object pose tracking datasets since it is the first to simulate the common scenario in which the camera and the objects are moving simultaneously under challenging conditions.

The rest of the article is structured as follows: Section~\ref{sec:tclch} presents the derivation of the cost function as well as the involved statistical segmentation model based on tclc-histograms. The systematic derivation of the Gauss-Newton optimization for pose estimation of this cost function is given in Section~\ref{sec:optimization}. This is followed by implementation details in Section~\ref{sec:implementation} and the introduction of our dataset in Section~\ref{sec:eval}, where also an extensive experimental evaluation of our approach is provided. The article concludes with Section~\ref{sec:conc} and the acknowledgements in Section~\ref{sec:ack}.
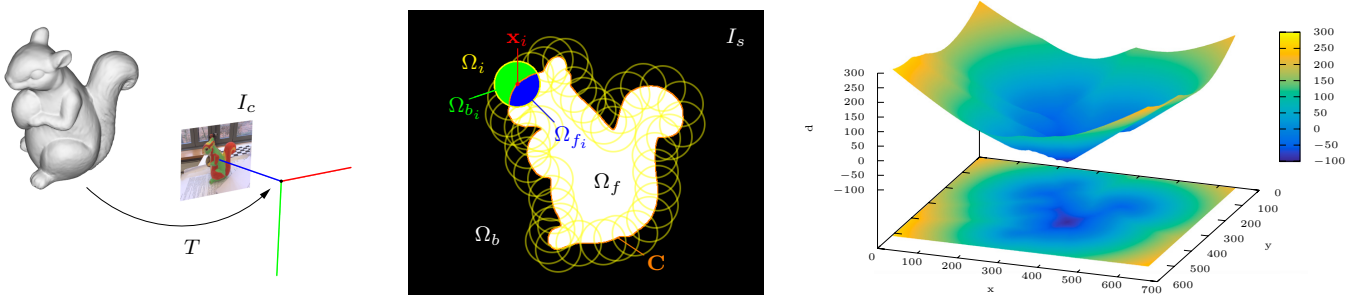
\begin{figure*}[!htp]
\centering
\begin{picture}(300,110)

\put(0,0){\includegraphics[width=0.26\textwidth]{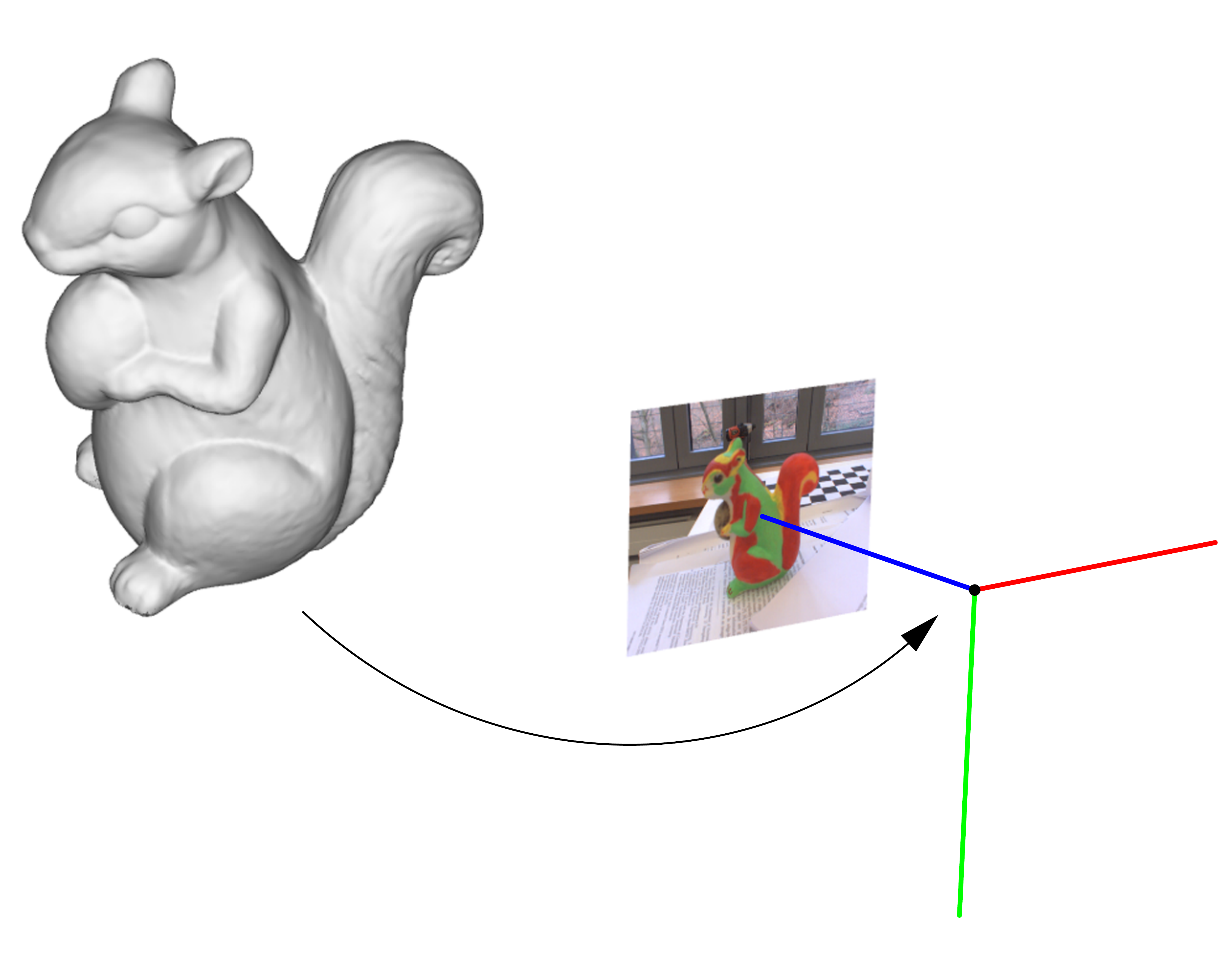}\hfill}

\put(155,0){\includegraphics[width=0.26\textwidth]{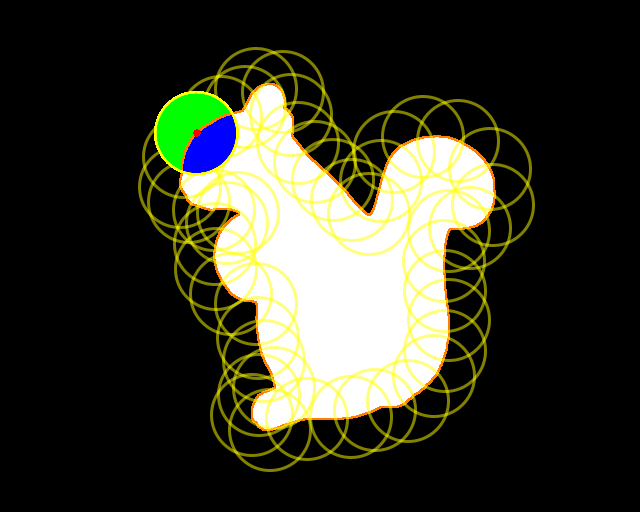}\hfill}

\begin{footnotesize}

\put(70, 15){\textcolor{black}{$T$}}
\put(90, 70){\textcolor{black}{$I_c$}}
\put(275, 95){\textcolor{white}{$I_s$}}

\put(225, 40){\textcolor{black}{$\Omega_f$}}
\put(180, 20){\textcolor{white}{$\Omega_b$}}

\put(234, 21.4){\color{orange}\line(3, -2){10}}
\put(245, 9){\textcolor{orange}{$\mathbf{C}$}}

\put(175, 85){\textcolor{yellow}{$\Omega_i$}}

\put(190, 78){\color{green}\line(-3, -1){12}}
\put(170, 69){\textcolor{green}{$\Omega_{b_i}$}}

\put(200, 75){\color{blue}\line(1, -1){10}}
\put(210, 58){\textcolor{blue}{$\Omega_{f_i}$}}

\put(196.2, 80){\color{red}\line(0, 1){13}}
\put(192, 95){\textcolor{red}{$\mathbf{x}_i$}}

\end{footnotesize}

\end{picture}
\includegraphics[width=0.4\textwidth]{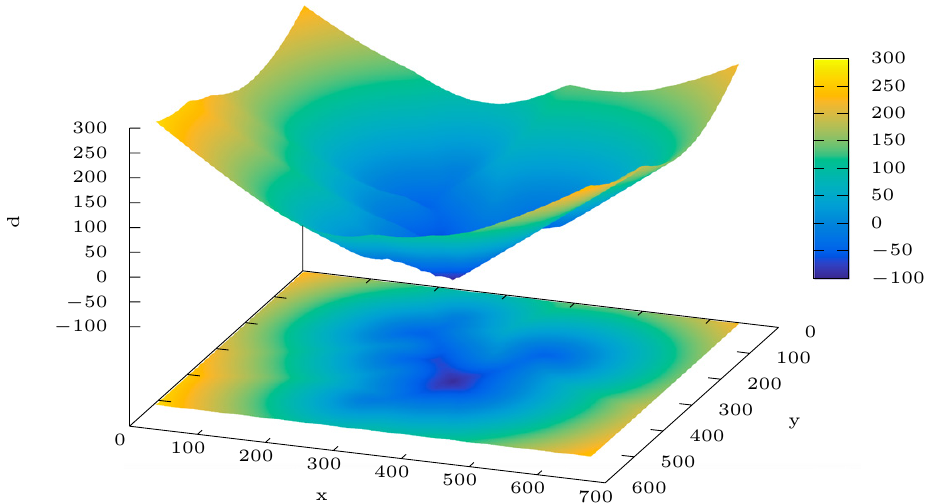}
\caption{An overview of our region-based pose estimation setting for a single object. Left: The object pose $T$ relative to a camera based on a color image $I_c$ and a 3D model of the object. Middle: The object's silhouette $I_s$ generated by projecting the 3D surface model into the 2D image plane using an estimated pose $T$. Right: A combined 2D/3D plot of the level-set pose embedding $\Phi(\mathbf{x})$ in form of an Euclidean signed distance transform of the projected silhouette.}
\label{fig:overview}
\end{figure*}

\section{Temporally Consistent Local Color Histograms for Pose Tracking} \label{sec:tclch}

We begin by giving an overview of basic mathematical concepts and the notation used within this work. Then we will derive the region-based cost function based on tclc-histograms. To simplify notation, we first consider the case of tracking a single object and then extend it to multiple objects.

\subsection{Preliminaries}
 In this work we represent each object by a 3D model in form of a triangle mesh with vertices $\mathbf{X}_i = [X_i, Y_i, Z_i]^\top \in \mathbb{R}^3$, $i = 1,\dots, n$. We denote a camera RGB color image by $I_c : \Omega \subset \mathbb{R}^2 \rightarrow \{0, \dots, 255\}^3$ assuming 8-bit quantization per intensity channel. The color at a pixel location $\mathbf{x} = [x,y]^\top \in \mathbb{R}^2$ in the 2D image plane is then given by $\mathbf{y} = I_c(\mathbf{x})$. By projecting a 3D model into the image plane we obtain a binary silhouette mask denoted by $I_s : \Omega \subset \mathbb{R}^2 \rightarrow \{0,1\}$ that yields a contour $\mathbf{C}$ splitting the image into a foreground region $\Omega_f \subset \Omega$ and a background region $\Omega_b = \Omega \setminus \Omega_f$ (see Figure~\ref{fig:overview}). 
 
 The pose of an object describing the rigid body transform from its 3D model coordinate frame to the camera coordinate frame is represented by a $4 \times 4$ homogeneous matrix
\begin{equation}
T = 
\begin{bmatrix}
R & \mathbf{t} \\
\mathbf{0} & 1
\end{bmatrix}
\in \mathbb{SE}(3),
\text{ with }
R \in \mathbb{SO}(3)
\text{ and }
\mathbf{t} \in \mathbb{R}^3
\end{equation}
being an element of the Lie-group $\mathbb{SE}(3)$. We assume the intrinsic  parameters of the camera to be known from an offline pre-calibration step. By denoting the linear projection parameters in form of a $3 \times 3$ matrix
\begin{equation}
K = 
\begin{bmatrix}
f_x & 0 & c_x \\
0 & f_y & c_y \\
0 & 0 & 1
\end{bmatrix}
\in \mathbb{R}^{3 \times 3},
\label{eq:K}
\end{equation}
and assuming that all images have been rectified by removing lens distortion, we describe the projection of a 3D model surface point $\mathbf{X}$ into the image plane by
\begin{equation}
\mathbf{x} = \pi\bigl(K(T\mathbf{\tilde X})_{3 \times 1}\bigr),
\end{equation}
with $\pi(\mathbf{X}) = [X/Z, Y/Z]^\top$. Here, the \textit{tilde-notation} marks the homogeneous representation $\mathbf{\tilde X} = [X,Y,Z,1]^\top$ of the point $\mathbf{X} = [X,Y,Z]^\top = (\mathbf{\tilde X})_{3 \times 1}$.

For pose tracking we denote a time-discrete sequence of images by $I_c(t_k)$. Each image is captured at time $t_k \in \mathbb{R}$, $k = 0, \dots, l$, with $I_c(t_l)$ being the current live image. Accordingly, we compute the trajectory of an object by estimating a sequence of rigid body transformations $T(t_k)$, $k = 0, \dots, l$, each corresponding to the related video frame. By assuming that the pose $T(t_{l-1})$ in the previous frame $I_c(t_{l-1})$ is known, we perform pose tracking in form of a so-called \textit{recursive pose estimation}. For this we express the current live pose as $T(t_l) = \Delta T T(t_{l-1})$. Here $\Delta T = T(t_l)T^{-1}(t_{l-1})$ is the pose difference that occurred between the last and the current frame. For a new live image $I_c(t_l)$ we thus always only need to compute the remaining $\Delta T$ in order to obtain the current live pose $T(t_l)$, as long we do not lose tracking. 

For pose optimization, we model the rigid body motion $\Delta T$ between $I_c(t_{l-1})$ and $I_c(t_l)$ with twists 
\begin{equation}
\hat \xi = 
\begin{bmatrix}
\mathbf{\hat w} & \mathbf{v} \\
\mathbf{0} & 0
\end{bmatrix}
\in \mathfrak{se}(3),
\text{ with }
\mathbf{\hat w} \in \mathfrak{so}(3)
\text{ and }
\mathbf{v} \in \mathbb{R}^3
\end{equation}
being elements of the  Lie-algebra $\mathfrak{se}(3)$ corresponding to the Lie-group $\mathbb{SE}(3)$. Each twist is parametrized by a six-dimensional vector of so-called \textit{twist coordinates}
\begin{equation}
\xi = 
\begin{bmatrix}
\mathbf{w} \\ \mathbf{v}
\end{bmatrix}
= [\omega_1, \omega_2, \omega_3, v_1, v_2, v_3]^\top
\in \mathbb{R}^6,
\end{equation}
and the matrix exponential
\begin{equation}
\Delta T = \exp(\hat \xi) 
\in \mathbb{SE}(3)
\end{equation}
maps a twist to its corresponding rigid body transformation. For detailed information on Lie groups and Lie algebra please refer to e.g.~\cite{ma05}.

\subsection{The Region-based Cost Function}

Our approach is essentially based on statistical image segmentation~\cite{cremers07}. As usual in this context, we represent the object's silhouette implicitly by a so-called \textit{shape-kernel} $\Phi(\mathbf{x})$. This is a level-set embedding of the object's shape such that the zero-level line $\mathbf{C} = \{\mathbf{x} \;|\; \Phi(\mathbf{x}) = 0\}$ gives its contour, i.e. the boundary between $\Omega_f $ and $\Omega_b$. Here, we use the shape-kernel
\begin{equation}
\Phi(\mathbf{x}) = \begin{cases}
     -d(\mathbf{x}, \mathbf{C}) & \forall  \mathbf{x} \in  \Omega_f\\
     \hphantom{-}d(\mathbf{x}, \mathbf{C}) & \forall  \mathbf{x} \in  \Omega_b 
   \end{cases}, 
   \label{eq:sdt}
\end{equation}
with
\begin{equation}
d(\mathbf{x}, \mathbf{C}) = \min_{\mathbf{c} \in \mathbf{C}}\|\mathbf{c} - \mathbf{x}\|_2,
\end{equation}
being the Euclidean distance between a pixel position $\mathbf{x}$ and the contour of the binary silhouette mask $I_s$.
 
For shape matching and segmentation we adopt the probabilistic formulation, as originally derived in~\cite{bibby08} and later used within PWP3D~\cite{prisacariu12}:
\begin{equation}
\begin{aligned}
P(\Phi|I_c) = \prod_{\mathbf{x} \in \Omega} \bigl(&H_e(\Phi(\mathbf{x}))P_f(\mathbf{x}) \\ 
&+ (1 - H_e(\Phi(\mathbf{x})))P_b(\mathbf{x}) \bigr).
\end{aligned}
\label{eq:probabilisiticformulation}
\end{equation}
It describes the posterior probability of the shape kernel $\Phi$ given an image $I_c$, with $H_e$ being a smoothed Heaviside step function. Here, $P_f(\mathbf{x})$ and $P_b(\mathbf{x})$ represent the per pixel foreground and background region membership probability, based on the underlying statistical appearance models (see Section~\ref{sec:segmentationmodel}). In the general context of 2D region-based image segmentation, the closed curve $\mathbf{C}$ would be evolved in an unconstrained manner such that it maximizes $P(\Phi|I_c)$ and thus the discrepancy between the foreground and background appearance model statistics. In our scenario, however, the evolution of the objects contour $\mathbf{C}$ is constrained by the known shape prior in form of a 3D model. Therefore, the shape kernel only depends on the pose parameters, i.e. $\Phi(\mathbf{x}(\xi))$. Assuming pixel-wise independence and taking the negative log of~\eqref{eq:probabilisiticformulation}, we obtain the region-based cost function
\begin{equation}
\begin{aligned}
E(\xi) = -\sum_{\mathbf{x} \in \Omega} \log\bigl(&H_e(\Phi(\mathbf{x}(\xi)))P_f(\mathbf{x})\\ 
&+ (1 - H_e(\Phi(\mathbf{x}(\xi))))P_b(\mathbf{x}) \bigr).
\end{aligned}
\label{eq:energy}
\end{equation}
This function can be optimized with respect to twist coordinates $\xi$ for pose estimation based on 2D-to-3D shape matching. In our approach we define the Heaviside function explicitly as
\begin{equation}
H_e\left(\Phi(\mathbf{x})\right) = \frac{1}{\pi}\left(-\atan(s\cdot \Phi(\mathbf{x}))+\frac{\pi}{2}\right),
\label{eq:heaviside}
\end{equation}
with $s$ determining the pitch of the smoothed transition (see Section~\ref{sec:implementation} for details). 

\subsection{Statistical Segmentation Model} \label{sec:segmentationmodel}


\begin{figure}[!tp]
\centering
\begin{picture}(127,100)
\put(0,0){\includegraphics[width=0.496\columnwidth]{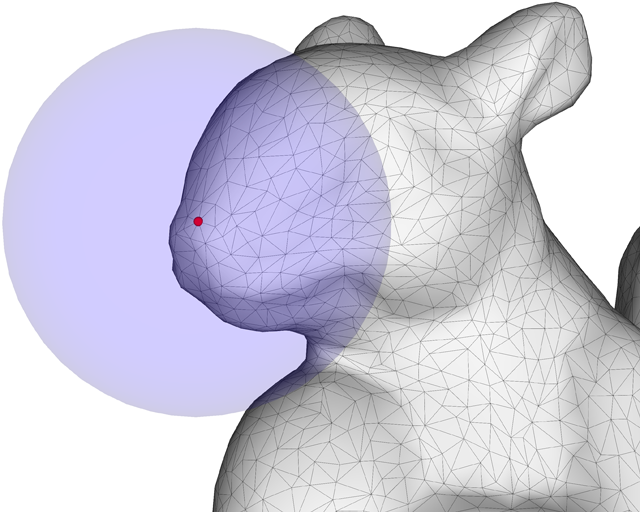}}
\footnotesize

\put(28,59){$\mathbf{X}_i$}

\end{picture}\hfill
\begin{picture}(125,100)
\put(0,0){\includegraphics[width=0.496\columnwidth]{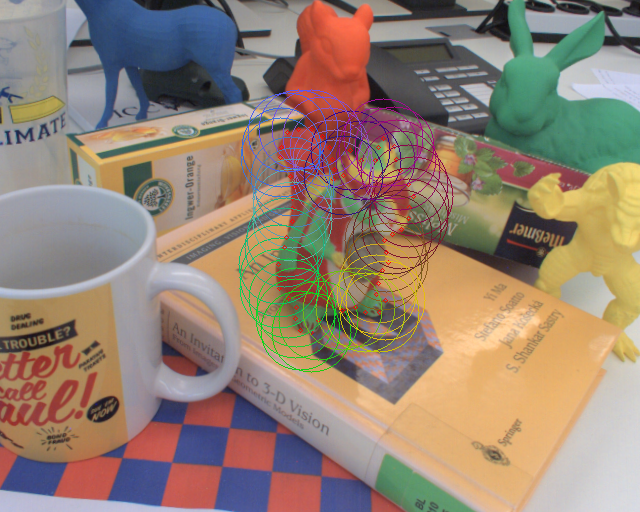}}
\footnotesize

\put(4,90){\textcolor{white}{$I_c$}}

\end{picture}\hfill\\[2pt]
\begin{picture}(127,100)
\put(0,0){\includegraphics[width=0.496\columnwidth]{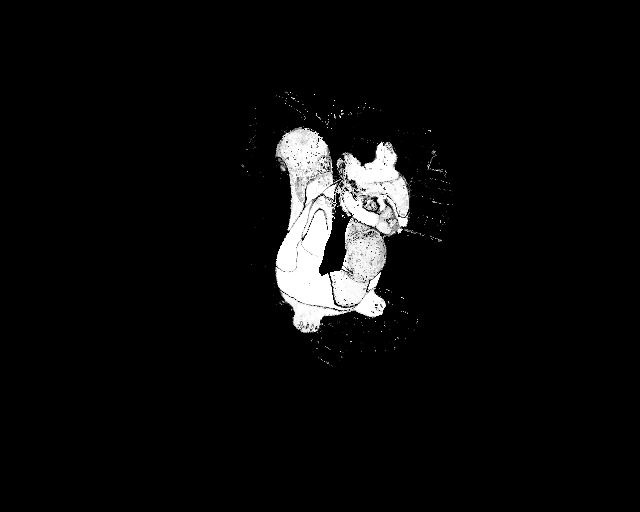}}
\footnotesize

\put(4,90){\textcolor{white}{$\bar P_f(\mathbf{x}) - \bar P_b(\mathbf{x})$}}

\end{picture}\hfill
\begin{picture}(125,100)
\put(0,0){\includegraphics[width=0.496\columnwidth]{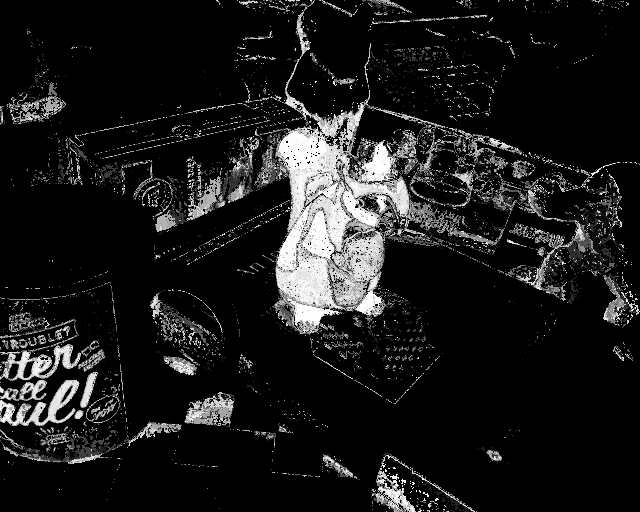}}
\footnotesize

\put(4,90){\textcolor{white}{$P_f(\mathbf{x}) - P_b(\mathbf{x})$}}

\end{picture}
\caption{Object segmentation using tclc-histograms. Top left: Schematic 3D visualization of a tclc-histogram attached to a mesh vertex $\mathbf{X}_i$ of a 3D squirrel model. Top right: A color image $I_c$ of a heterogeneous squirrel in a cluttered scene overlayed with the local regions. These are depicted by colored circles where the RGB value relates to the coordinates of the corresponding $\mathbf{X}_i$. Bottom left: Per pixel segmentation (visualized as $\bar P_f(\mathbf{x}) - \bar P_b(\mathbf{x}) > 0$) computed from the tclc-histograms~\eqref{eq:pbar}. Bottom right: Segmentation result from global color histograms~\eqref{eq:globalp} for comparison.}
\label{fig:segmentation}
\end{figure}
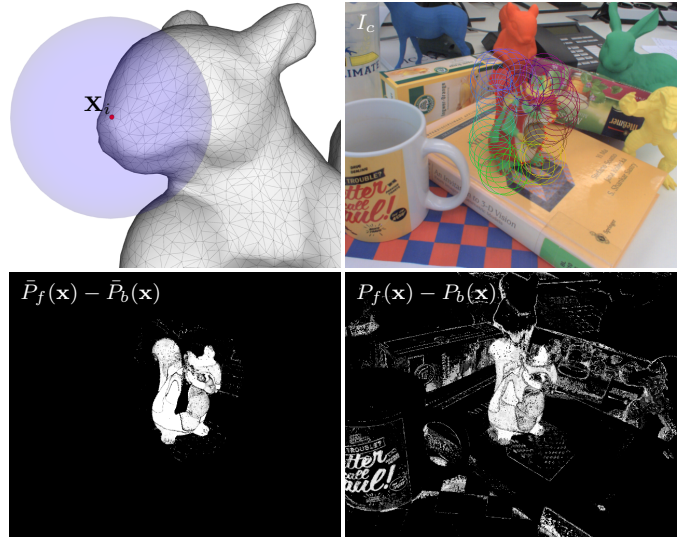

In the past, different appearance models have been proposed to compute $P_f(\mathbf{x})$ and $P_b(\mathbf{x})$ used in~\eqref{eq:energy}. Initially~\cite{bibby08} used a global appearance model based on the color distribution in both $\Omega_f$ and $\Omega_b$. Here each region has its own model denoted by $P(\mathbf{y} | M_f)$ for the foreground, i.e. the object and $P(\mathbf{y} | M_b)$ for the background. Each of them is represented with a global color histogram. Based on this, the region membership probabilities are calculated in form of pixel-wise posteriors as
\begin{eqnarray}
\begin{aligned}
P_f(\mathbf{x}) &= P(M_f | \mathbf{y}) =  \frac{P(\mathbf{y} | M_f)}{\eta_f P(\mathbf{y} | M_f) + \eta_b P(\mathbf{y} | M_b)},\\[1mm]
P_b(\mathbf{x}) &= P(M_b | \mathbf{y}) =  \frac{P(\mathbf{y} | M_b)}{\eta_f P(\mathbf{y} | M_f) + \eta_b P(\mathbf{y} | M_b)},
\end{aligned}
\label{eq:globalp}
\end{eqnarray}
where $\mathbf{y} = I_{c}(\mathbf{x})$ and
\begin{equation}
\eta_f = \sum_{\mathbf{x} \in \Omega} H_e(\Phi(\mathbf{x})), \quad \eta_b = \sum_{\mathbf{x} \in \Omega} 1 - H_e(\Phi(\mathbf{x})).
\end{equation}
This model is also used within PWP3D~\cite{prisacariu12} where it is further suggested to keep the appearance models temporally consistent, in order to adapt to scene changes while tracking the object. Having successfully estimated the current live pose $T(t_l)$ allows to render a corresponding silhouette mask denoted by $I_s(t_l)$. Ideally, this mask provides an exact segmentation of the object region in the current camera frame $I_c(t_l)$ and can thus be used in order to compute up-to-date color histograms $P^{t_l}(\mathbf{y} | M_f)$ and $P^{t_l}(\mathbf{y} | M_b)$. Instead of always using the latest color distribution for the appearance models,~\cite{prisacariu12} suggested to recursively adjust the histograms by
\begin{eqnarray}
\begin{aligned}
P(\mathbf{y} | M_f) &= (1-\alpha_f)P^{t_{l-1}}(\mathbf{y} | M_f) + \alpha_f P^{t_l}(\mathbf{y} | M_f),\\[1mm]
P(\mathbf{y} | M_b) &= (1-\alpha_b)P^{t_{l-1}}(\mathbf{y} | M_b) + \alpha_b P^{t_l}(\mathbf{y} | M_b),
\end{aligned}
\label{eq:globalupdate}
\end{eqnarray}
to prevent them from being corrupted by occlusions or pose estimation inaccuracies. Here, $\alpha_f$ and $\alpha_b$ denote foreground and background learning rates.

All the above is based on the assumption that the global color distribution is sufficiently descriptive in order to distinguish between the foreground and the background region. Therefore, this appearance model has been shown to work particularly well with homogeneous objects of a distinct color that is not dominantly present in the rest of the scene. However, for objects with heterogeneous surfaces and in case of cluttered scenes this global model is prone to fail.  

Hence, in~\cite{hexner16} a localized appearance model was proposed for the PWP3D approach that better captures spatial variations of the object's surface. The idea is to build the segmentation model from multiple overlapping circular image regions along the object's contour as originally introduced in~\cite{lankton08}. We denote each such local region by $\Omega_i = \{ \mathbf{x} \; \big| \; \|\mathbf{x} - \mathbf{x}_i\|_2 < r \}$ with radius $r$, centered at pixel $\mathbf{x}_i \in \mathbf{C}$. Now, $I_s$ splits each $\Omega_i$ into a foreground region $\Omega_{f_i} \subset \Omega_i$ and a background region $\Omega_{b_i} = \Omega_i \setminus \Omega_{f_i}$ (see Figure~\ref{fig:overview}). This allows to compute local foreground and background color histograms for each region. In~\cite{hexner16} this led to the localized cost function
\begin{equation}
\begin{aligned}
E = -\frac{1}{n} \sum_{i=1}^n \sum_{\mathbf{x} \in \Omega}&\log\bigl(H_e(\Phi(\mathbf{x}))P_{f_i}(\mathbf{x}) \\ &+ (1 - H_e(\Phi(\mathbf{x})))P_{b_i}(\mathbf{x})\bigr)\mathbf{B}_i(\mathbf{x}),
\label{eq:e_hexner}
\end{aligned}
\end{equation}
using the masking function 
\begin{equation}
\mathbf{B}_i(\mathbf{x}) =  
\begin{cases} 
1 & \forall \mathbf{x} \in \Omega_i \\ 
0 & \forall \mathbf{x} \not \in \Omega_i 
\end{cases},
\end{equation}
which indicates whether a pixel $\mathbf{x}$ lies within a local region or not. Here, the local region membership probabilities $P_{f_i}(\mathbf{x})$ and $P_{b_i}(\mathbf{x})$ are computed individually from the local histograms as
\begin{eqnarray}
\begin{aligned}
P_{f_i}(\mathbf{x}) &= \frac{P(\mathbf{y} | M_{f_i})}{\eta_{f_i} P(\mathbf{y} | M_{f_i}) + \eta_{b_i} P(\mathbf{y} | M_{b_i})},\\[1mm]
P_{b_i}(\mathbf{x}) &= \frac{P(\mathbf{y} | M_{b_i})}{\eta_{f_i} P(\mathbf{y} | M_{f_i}) + \eta_{b_i} P(\mathbf{y} | M_{b_i})},
\end{aligned}
\label{eq:pb}
\end{eqnarray}
where $\mathbf{y} = I_{c}(\mathbf{x})$ and
\begin{equation}
\eta_{f_i} = \sum_{\mathbf{x} \in \Omega_i} H_e(\Phi(\mathbf{x})), \quad \eta_{b_i} = \sum_{\mathbf{x} \in \Omega_i} 1 - H_e(\Phi(\mathbf{x})),
\end{equation}
in analogy to the global model. In~\cite{hexner16}, however, temporal consistency of the local appearance models was not addressed. The local region centers $\mathbf{x}_i$ were calculated as arbitrary sets of pixel locations along $\mathbf{C}$ for each image. Thus, this approach in general does not allow to establish correspondences of the centers across multiple frames (i.e.~$\mathbf{x}_i(t_l) \leftrightarrow \mathbf{x}_i(t_{l-1})$) which is required in order to update the respective histograms.

This issue has been addressed in~\cite{tjaden17} by introducing a segmentation model based on temporally consistent local color histograms (tclc-histograms), which we adopt in this work. Here, each 3D model vertex $\mathbf{X}_i$ is associated with a local foreground and background histogram (see Figure~\ref{fig:segmentation}). In contrast to~\cite{hexner16} this allows us to compute the histogram centers by projecting all model vertices into the image plane, i.e.~$\mathbf{x}_i = \pi\bigl(K(T\mathbf{\tilde X}_i)_{3\times1}\bigr)$ and selecting the subset of all $\mathbf{x}_i \in \mathbf{C}$. Since each histogram is anchored to the objects surface, center correspondences $\mathbf{x}_i(t_l) \leftrightarrow \mathbf{x}_i(t_{l-1})$  between frames are simply given by the projection of corresponding surface points, i.e.~$\pi\bigl(K(T(t_l)\mathbf{\tilde  X}_i)_{3\times1}\bigr) \leftrightarrow \pi\bigl(K(T(t_{l-1})\mathbf{\tilde X}_i)_{3\times1}\bigr)$. This ensures to keep the individual histograms temporally consistent. Whenever a model vertex projects onto the contour for the first time, its corresponding histograms are initialized from the local region around its center in the current frame. Otherwise, if its histograms already contain information from a previous frame, we update them as
\begin{eqnarray}
\begin{aligned}
P(\mathbf{y} | M_{f_i}) &= (1-\alpha_f)P^{t_{l-1}}(\mathbf{y} | M_{f_i}) + \alpha_f P^{t_l}(\mathbf{y} | M_{f_i}),\\[1mm]
P(\mathbf{y} | M_{b_i}) &= (1-\alpha_b)P^{t_{l-1}}(\mathbf{y} | M_{b_i}) + \alpha_b P^{t_l}(\mathbf{y} | M_{b_i}),
\end{aligned}
\end{eqnarray}
in analogy to~\eqref{eq:globalupdate}.

In~\cite{tjaden17} it has furthermore been shown that computing the average energy~\eqref{eq:e_hexner} over all local regions $\Omega_i$ potentially suffers from the same segmentation problems locally, as the previous approach based on the global appearance model. More robust results can be obtained by  computing the average posteriors from all local histograms instead as
\begin{eqnarray}
\begin{aligned}
\bar{P_f}(\mathbf{x}) &= \frac{1}{\sum_{i=1}^n \mathbf{B}_i(\mathbf{x})}\sum_{i=1}^n P_{f_i}(\mathbf{x})\mathbf{B}_i(\mathbf{x}),\\[1mm]
\bar{P_b}(\mathbf{x}) &= \frac{1}{\sum_{i=1}^n \mathbf{B}_i(\mathbf{x})}\sum_{i=1}^n P_{b_i}(\mathbf{x})\mathbf{B}_i(\mathbf{x}),
\end{aligned}
\label{eq:pbar}
\end{eqnarray}
and use these within~\eqref{eq:energy}. We now can define the energy function
\begin{equation}
\begin{aligned}
E(\xi) = -\sum_{\mathbf{x} \in \Omega} \log\bigl(&H_e(\Phi(\mathbf{x}(\xi)))\bar P_f(\mathbf{x})\\ 
&+ (1 - H_e(\Phi(\mathbf{x}(\xi))))\bar P_b(\mathbf{x}) \bigr),
\end{aligned}
\label{eq:final_energy}
\end{equation}
that we use for our pose tracking approach based on tclc-histograms.

\subsection{Extension to Using Multiple Objects}

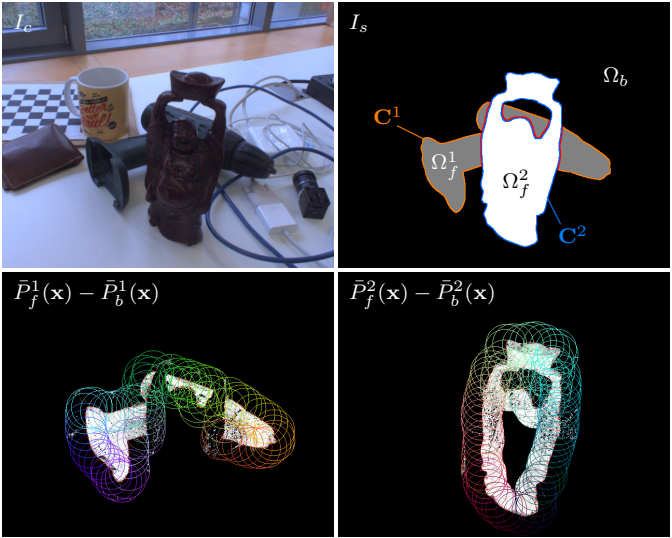
\begin{figure}[!tp]
\centering
\begin{picture}(127,100)
\put(0,0){\includegraphics[width=0.496\columnwidth]{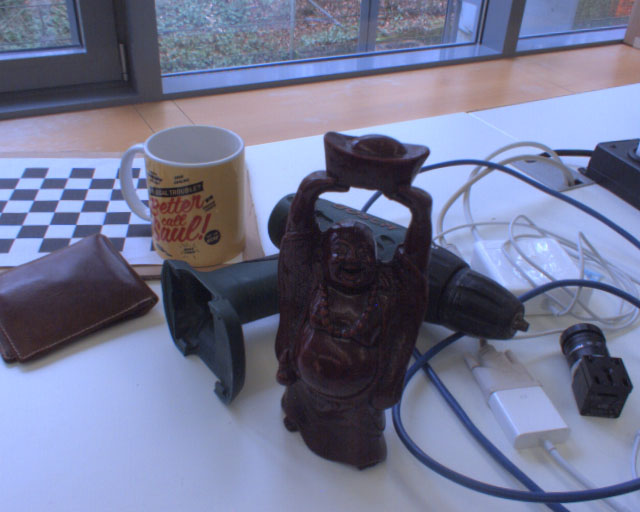}}
\footnotesize

\put(4,90){\textcolor{white}{$I_c$}}

\end{picture}\hfill
\begin{picture}(125,100)
\put(0,0){\includegraphics[width=0.496\columnwidth]{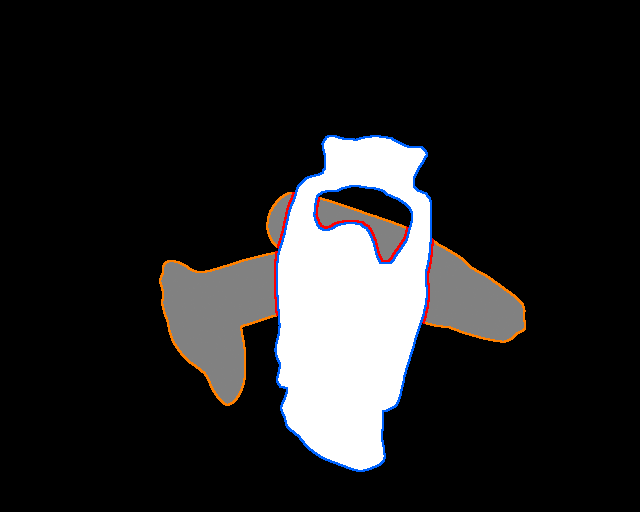}}
\footnotesize

\put(4,90){\textcolor{white}{$I_s$}}

\put(35,37){\textcolor{white}{$\Omega_f^1$}}
\put(62,30){$\Omega_f^2$}
\put(100,70){\textcolor{white}{$\Omega_b$}}

\put(22.2,53.5){\color{orange}\line(2, -1){10}}
\put(13,54){\textcolor{orange}{$\mathbf{C}^1$}}

\put(84.0,16.3){\color{myblue}\line(-1, 2){5}}
\put(83,9){\textcolor{myblue}{$\mathbf{C}^2$}}

\end{picture}\hfill\\[2pt]
\begin{picture}(127,100)
\put(0,0){\includegraphics[width=0.496\columnwidth]{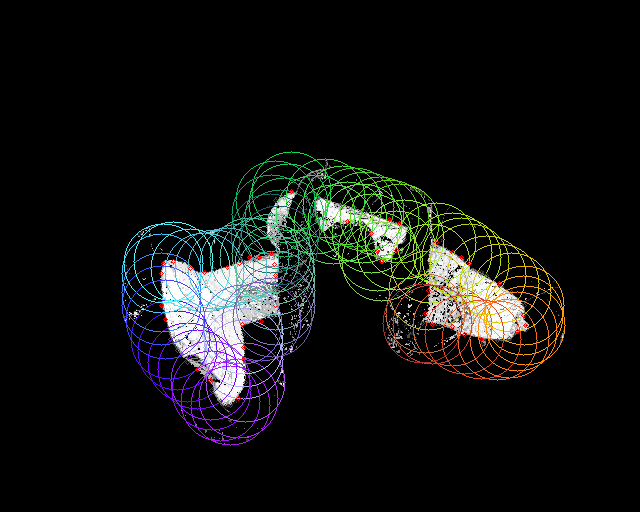}}
\footnotesize

\put(4,90){\textcolor{white}{$\bar P_f^1(\mathbf{x}) - \bar P_b^1(\mathbf{x})$}}

\end{picture}\hfill
\begin{picture}(125,100)
\put(0,0){\includegraphics[width=0.496\columnwidth]{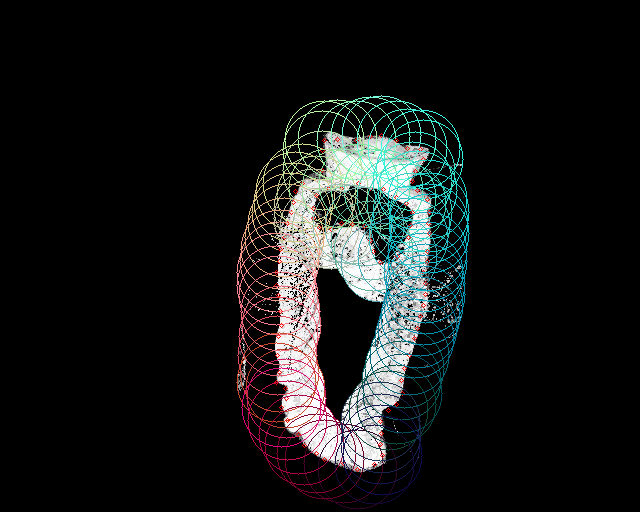}}
\footnotesize

\put(4,90){\textcolor{white}{$\bar P_f^2(\mathbf{x}) - \bar P_b^2(\mathbf{x})$}}

\end{picture}
\caption{A multi-object tracking scenario. Top left: Image $I_c$ of a driller behind a Buddha figurine. Top right: Estimated common silhouette mask $I_s$, where the segments of $\mathbf{C}^1$, that appear due to the occlusion, are marked red. Bottom: Corresponding per pixel segmentation computed from the tclc-histograms of each object. This shows that even in case of these rather dark, mutually occluding objects, the segmentation strategy produces high quality results.
}
\label{fig:multimask}
\end{figure}

In the following, we will explain how our approach easily extends to tracking multiple objects simultaneously, similar to~\cite{prisacariu12}. For this each object is represented by its own 3D model, with corner vertices $\mathbf{X}_i^j$, $j = 1, \dots, m$, where $m$ is the total number of objects. Accordingly, the individual poses are denoted by $T^j$. Projecting all models into the image plane yields a common segmentation mask $I_s : \Omega \subset \mathbb{R}^2 \rightarrow \{1, \dots, m\}$. It contains $m$ contours $\mathbf{C}^j$ that split the image into multiple foreground regions $\Omega_f^j \subset \Omega$ and background regions $\Omega_b^j = \Omega \setminus \Omega^j_f$ (see Figure~\ref{fig:multimask}).

For pose tracking we optimize a separate energy function for each object denoted by
\begin{equation}
\begin{aligned}
E^j(\xi^j) = -\sum_{\mathbf{x} \in \Omega} \log\bigl(&H_e(\Phi^j(\mathbf{x})(\xi^j))\bar P_f^j(\mathbf{x})\\ 
&+ (1 - H_e(\Phi^j(\mathbf{x})))\bar P_b^j(\mathbf{x}) \bigr),
\end{aligned}
\end{equation}
with its own level-set $\Phi^j$ and region membership probabilities $\bar P_f^j(\mathbf{x})$ and $\bar P_b^j(\mathbf{x})$ computed from its individual set of $n^j$ tclc-histograms. Here, $n^j$ denotes the number of vertices per model. Each such optimization is performed regardless of the other objects as long as they do not occlude each other. However, in cases of mutual occlusions, the foreground regions overlap, which results in contour segments that do not belong to the actual silhouette of the objects (see again Figure~\ref{fig:multimask}). These cases must be detected and handled appropriately during pose optimization as explained in detail in Section~\ref{sec:occlusions}.  

This shows that for all formulas, extending our approach to multiple objects is essentially done by adding the object index $j$. For the sake of clarity we will drop $j$ in the rest of this article again, unless absolutely required.

\section{Pose Optimization} \label{sec:optimization}

Traditionally, cost functions of form~\eqref{eq:energy} have been optimized using gradient descent (see e.g.~\cite{prisacariu12} or~\cite{hexner16}).  In comparison to second-order (Newton) methods, this has several drawbacks. First, one has to determine suitable time step sizes associated with translation and rotation. Too small step sizes often lead to very slow convergence, too large step sizes easily induce oscillations and instabilities. For the gradient descent-based PWP3D method \cite{prisacariu12}, for example, one needs to manually specify the number of iterations and three different step sizes (one each for rotation, translation along the optical axis and translation within the camera€™s image plane).  These need to be adapted at least once for each new object.  Moreover, as shown in the exemplary comparison in Figure \ref{fig:pwp3d}, often no suitable compromise between numerical stability and convergence to the desired solution can be achieved. Second, convergence is typically not as robust and rather slow (especially near the optimum), making the technique less suitable for accurate and robust real-time tracking. 

Applying a second-order optimization scheme, on the other hand, is not straightforward because the cost function \eqref{eq:energy} is not in the classical form of a nonlinear least-squares problem. In the following, we will propose a strategy to circumvent this issue, based on rewriting the original problem in form of a re-weighted nonlinear least-squares estimation. This allows us to apply a Gauss-Newton algorithm, being also the method of choice for state-of-the-art real-time visual SLAM methods, such as Direct Sparse Odometry~\cite{engel18}. Our approach is different from an (in our view less straightforward) derivation proposed by Bibby and Reid \cite{bibby08}, which requires a Taylor series approximation of the square root. Finally, strategies involving line searches (e.g. Levenberg-Marquardt) are not suitable for real-time tracking, as they would require many evaluations of~\eqref{eq:energy}, which are almost as costly as calculating its derivatives. Both require computing $I_s$ and $\Phi(\mathbf{x})$.

\subsection{Derivation of a Gauss-Newton Strategy} \label{sec:gn}

The cost function in  \eqref{eq:final_energy} can be written compactly as
\begin{equation}
\begin{aligned}
E(\xi) =& \sum_{\mathbf{x} \in \Omega} F(\mathbf{x}, \xi), \quad \text{where}\\
F(\mathbf{x}, \xi) =& -\log\bigl(H_e(\Phi(\mathbf{x}(\xi)))\bar{P_f}(\mathbf{x}) \\&\hphantom{---}+ (1 - H_e(\Phi(\mathbf{x}(\xi))))\bar{P_b}(\mathbf{x}) \bigr).
\end{aligned}
\end{equation}
Unfortunately, this is not in the traditional form of a nonlinear least-squares estimation problem for which the Gauss-Newton algorithm is applicable.  However, we can simply rewrite this expression as a \textit{nonlinear weighted least-squares problem} of the form
\begin{equation}
E(\xi) =  \frac{1}{2}\sum_{\mathbf{x} \in \Omega} \psi(\mathbf{x})F^2(\mathbf{x}, \xi), \text{ with } \psi(\mathbf{x}) = \frac{1}{F(\mathbf{x}, \xi)}.
\end{equation}
To optimize this cost function, one can apply the technique of iteratively re-weighted least-squares estimation which amounts to solving the above problem for fixed weights $\psi(\mathbf{x})$ by means of Gauss-Newton optimization and alternatingly using the refined pose for updating the weights $\psi(\mathbf{x})$.  Over the iterations, these weights will adaptively re-weight respective terms.  

In the fixed-weight assumption, the gradient is given by
\begin{equation}
\frac{\partial E(\xi)}{\partial \xi} = \frac{1}{2}\sum_{\mathbf{x} \in \Omega}\psi(\mathbf{x})\frac{\partial F^2(\mathbf{x}, \xi)}{\partial \xi} = \sum_{\mathbf{x} \in \Omega}\psi(\mathbf{x})F\frac{\partial F}{\partial \xi},
\end{equation}
with $\psi(\mathbf{x})F(\mathbf{x}, \xi) = 1$, and the Hessian is given by
\begin{equation}
\frac{\partial^2 E(\xi)}{\partial \xi^2} =  \sum_{\mathbf{x} \in \Omega} \psi(\mathbf{x})\left(\left(\frac{\partial F}{\partial \xi}\right)^\top\frac{\partial F}{\partial \xi} + F\frac{\partial^2 F}{\partial \xi^2}\right).
\end{equation}
The Gauss-Newton algorithm emerges when applying a Newton method and dropping the second-order derivative of the residual $F$. This approximation is valid if either the residual itself is small (i.e. $F\approx 0$ near the optimum) or if the residuum is close to linear (in which case $\nicefrac{\partial^2 F}{\partial \xi^2}\approx 0$). If we denote the Jacobian of the residuum at the current pose $\xi$ by
\begin{equation}
J = \frac{\partial F(\mathbf{x}, \xi)}{\partial \xi},
\label{eq:jacobian}
\end{equation}
under the above assumptions, the second-order Taylor approximation of the cost function $E$ is given by
\begin{equation}
E(\xi + \Delta \xi) \approx E(\xi) + \sum_{\mathbf{x} \in \Omega} J \Delta \xi + \frac{1}{2} \sum_{\mathbf{x} \in \Omega} \psi(\mathbf{x})\Delta \xi^\top J^\top J \Delta \xi.
\label{gaussnewtonapprox}
\end{equation}
This leads to the optimal Gauss-Newton update step of
\begin{equation}
\Delta \xi = -\left(\sum_{\mathbf{x} \in \Omega} \psi(\mathbf{x})J^\top J\right)^{-1} \sum_{\mathbf{x} \in \Omega} J^\top.
\label{eq:updatestep}
\end{equation}
We apply this step as composition of the matrix exponential of the corresponding twist $\Delta\hat{\xi}$ with the previous pose as 
\begin{equation}
T \leftarrow \exp(\Delta\hat{\xi})T,
\end{equation}
in order to remain within the group $\mathbb{SE}(3)$.

\subsection{Computation of the Derivatives}

The per pixel Jacobian term~\eqref{eq:jacobian} is computed by applying the chain-rule as
\begin{equation}
\begin{aligned}
J = \frac{\bar P_b(\mathbf{x}) - \bar P_f(\mathbf{x})}{H_e(\Phi(\mathbf{x}))(\bar P_f(\mathbf{x})  - \bar P_b(\mathbf{x})) + \bar P_b(\mathbf{x})} \delta_e \frac{\partial \Phi(\mathbf{x}(\xi))}{\partial \xi},
\end{aligned}
\label{eq:j}
\end{equation}
where $\delta_e = \delta_e\left(\Phi(\mathbf{x})\right)$ is the smoothed Dirac delta function corresponding to $H_e$, i.e.
\begin{equation}
\delta_e\left(\Phi(\mathbf{x})\right) = \frac{s}{\pi \Phi(\mathbf{x})^2s^2 + \pi}.
\label{eq:dirac}
\end{equation}
Since $\mathbf{x}(\xi) = \pi\bigl(K(\exp(\hat \xi)T\mathbf{\tilde X})_{3 \times 1}\bigr)$, the derivatives of the signed distance transform are given by
\begin{equation}
\frac{\partial \Phi(\mathbf{x}(\xi))}{\partial \xi} = \frac{\partial \Phi}{\partial \mathbf{x}}\frac{\partial \pi}{\partial K(\exp(\hat \xi)T\mathbf{\tilde X})_{3 \times 1}}\frac{\partial K(\exp(\hat \xi)T\mathbf{\tilde X})_{3 \times 1}}{\partial \xi}.
\end{equation}
Assuming small motion, we perform piecewise linearization of the matrix exponential in each iteration, i.e.~$\exp(\hat \xi) \approx \mathbb{I}_{4\times4} + \hat \xi$ and therefore we get
\begin{equation}
\begin{aligned}
\frac{\partial \Phi(\mathbf{x}, \xi_0)}{\partial \xi} =& \begin{bmatrix} \frac{\partial \Phi}{\partial x}, \frac{\partial \Phi}{\partial y} \end{bmatrix} \begin{bmatrix} \frac{f_x}{Z'} & 0 & -\frac{X' f_x}{(Z')^2} \\[1mm] 0 & \frac{f_y}{Z'} & -\frac{Y' f_y}{(Z')^2} \end{bmatrix} \\
&\cdot\begin{bmatrix} 0 & Z' & -Y' & 1 & 0 & 0 \\-Z' & 0 & X' & 0 & 1 & 0 \\ Y' &-X' & 0 & 0 & 0 & 1 \end{bmatrix},
\end{aligned}
\label{eq:jacobianvector}
\end{equation}
with $\mathbf{X}' = [X', Y', Z']^\top = (T\tilde{\mathbf{X}})_{3 \times 1}$. For pixels $\mathbf{x} \in \Omega_b$, we choose $\mathbf{X}'$ to be the 3D surface point in the camera's frame of reference that projects to its closest contour pixel $\mathbf{c} \in \mathbf{C}$. Finally, we compute the derivatives of $\Phi(\mathbf{x})$ with respect to a pixel $\mathbf{x}$ as 2D image gradients, utilizing central differences, i.e. 
\begin{equation}
 \begin{bmatrix} \frac{\partial \Phi(\mathbf{x})}{\partial x}, \frac{\partial \Phi(\mathbf{x})}{\partial y} \end{bmatrix} = \begin{bmatrix}\nabla_x \Phi \\ \nabla_y \Phi \end{bmatrix}^\top = \begin{bmatrix}\frac{\Phi(x+1, y) - \Phi(x-1, y)}{2} \\  \frac{\Phi(x, y+1) - \Phi(x, y-1)}{2} \end{bmatrix}^\top.
\label{eq:sdt_deriv}
\end{equation}

In order to increase the convergence speed, iterative pose optimization is computed in an hierarchical coarse to fine approach. This also makes the tracking more robust towards fast movement or motion blur. Details on our concrete multi-level implementation are given in Section~\ref{sec:implopti}.

\subsection{Initialization}

During successful tracking, for every new frame $I_c(t_l)$ the optimization starts at the previously estimated pose $T(t_{l-1})$. Note that also the histogram centers $\mathbf{x}_i$, i.e. the regions $\Omega_i$ are obtained from $T(t_{l-1})$ and remain unchanged during the entire iterative optimization process. They provide the information used to compute $\bar P_f(\mathbf{x})$ and $\bar P_b(\mathbf{x})$ from the intensities in current frame.

To start tracking or recover tracking from tracking loss, our approach can be combined with a pose detection module in order to obtain the initial pose. As shown in~\cite{tjaden17}, with the help of manual initialization, tclc-histograms can act as an object descriptor for pose detection based on template matching. Due to the employed temporal consistency strategy this descriptor is trained online within a couple of seconds by tracking the object and showing it to the camera from different perspectives. This approach is particularly efficient to recover from temporary tracking loss, e.g. caused by massive occlusion or if the object leaves the camera's field of view. Another advantage of this approach is that it can be computed at frame rates of 4 -- 10\,Hz for a single object on commodity laptop CPU. However, the tclc-histogram based descriptors struggle in previously unseen environments if the foreground and background color distribution differs too much from the scene they were originally trained in.

When more computational power is available, recent deep learning-based approaches (see e.g.~\cite{crivellaro18, rad17, kehl17b}) also could be used for pose detection. These are currently achieving state-of-the-art results and can be trained only from synthetic images~\cite{kehl17b, hinterstoisser17} which make them robust to different environments and lighting conditions. However, they require a powerful GPU in order to achieve similar frame rates as our method based on tclc-histograms running on a CPU. 
\section{Implementation} \label{sec:implementation}

In the following we provide an overview of our C++ implementation with regard to runtime performance aspects. We perform all major processing steps in parallel on the CPU and only use the GPU via OpenGL for rendering purposes.

\subsection{Rendering Engine}

We use the standard rasterization pipeline of OpenGL in order obtain the silhouette masks $I_s$. Since we want to process the rendered images on the CPU, we perform offscreen rendering into a \textit{FrameBufferObject}, which is afterwards downloaded to host memory. To generate synthetic views that match the real images, the intrinsic parameters~\eqref{eq:K} of the camera need to be included. For this, we model the transformation from 3D model coordinates $\mathbf{X}$ to homogeneous coordinates within the canonical view volume of OpenGL as $\mathbf{\tilde V} = P(K)LT\mathbf{\tilde X} \in \mathbb{R}^4$. Here, $L$ is the so-called \textit{look-at} matrix
\begin{equation}
    L = 
    \begin{bmatrix}
    1 & 0 & 0 & 0 \\ 
    0 & -1 & 0 & 0 \\ 
    0 & 0 & -1 & 0 \\
    0 & 0 & 0 & 1
    \end{bmatrix} \in \mathbb{R}^{4\times 4},
\end{equation}
that aligns the principal axes of the real cameras coordinate frame with those of the virtual OpenGL camera and $P(K)$ is a homogeneous projection matrix
\begin{equation}
    P(K) = 
    \begin{bmatrix}
    \frac{2f_x}{w}  & 0 & 1 - \frac{2c_x}{w} & 0 \\ 
    0 & -\frac{2f_y}{h} & 1 - \frac{2c_y}{h} & 0 \\ 
    0 & 0 & -\frac{Z_f + Z_n}{Z_f - Z_n} & -\frac{2Z_fZ_n}{Z_f - Z_n} \\
    0 & 0 & -1 & 0
    \end{bmatrix} \in \mathbb{R}^{4\times 4},
\end{equation}
with respect to the camera matrix $K$. The scalars $w$, $h$ are the width and height of the real image $I_c$ and $Z_n$, $Z_f$ are the near- and far-plane of the view frustum described by $P(K)$.

\begin{figure}[!tp]
\centering
\begin{picture}(127,100)
\put(0,0){\includegraphics[width=0.496\columnwidth]{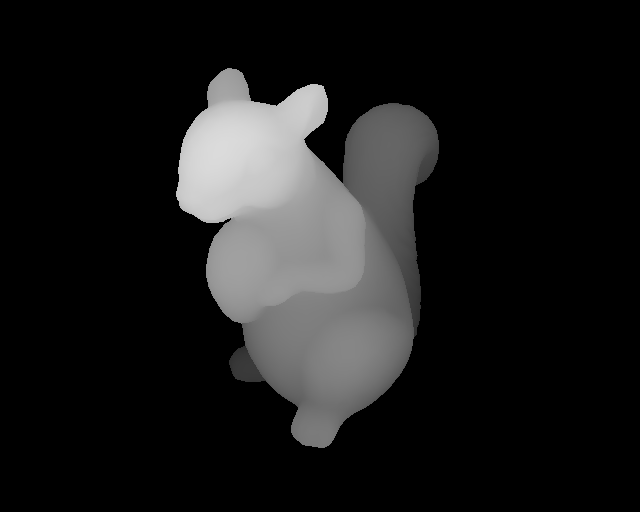}}
\footnotesize

\put(4,90){\textcolor{white}{$I_d$}}

\end{picture}\hfill
\begin{picture}(125,100)
\put(0,0){\includegraphics[width=0.496\columnwidth]{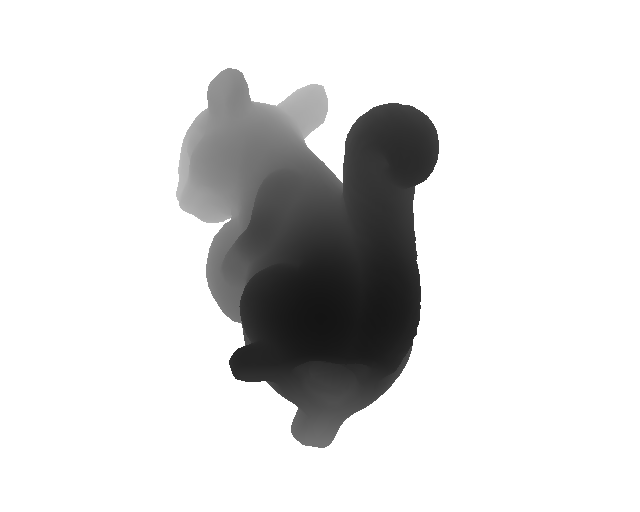}}
\footnotesize

\put(4,90){\textcolor{black}{$I_d^{r}$}}

\end{picture}
\caption{The two depth map types used within our approach, where brighter pixels are closer to the camera. Left: The usual depth map $I_d$ corresponding to the closest surface points. Right: The reverse depth map $I_d^{r}$ corresponding to the most distant surface points.}
\label{fig:depthmaps}
\end{figure}

In case of tracking multiple objects, all 3D models are rendered in the same scene. Each mesh is rendered with a constant and unique color that corresponds to its model index $j$. This allows to separate the individual foreground regions and identify their contours $\mathbf{C}^j$ as required for computing the different level-sets. Here, mutual occlusions are natively handled by the OpenGL $Z$-Buffer.

As seen in~\eqref{eq:jacobianvector}, the derivatives used for pose optimization involve the coordinates of the 3D surface point $\mathbf{X}'$ in the camera's frame of reference, corresponding to each pixel $\mathbf{x}$.
In addition to the silhouette mask, we therefore also download the $Z$-buffer into a per pixel depth map $I_d : \Omega \rightarrow [0,1] \subset \mathbb{R}$. Given $I_d$, the required coordinates are efficiently determined via backprojection as $\mathbf{X}'(\mathbf{x}, I_d) = D(\mathbf{x}, I_d)K^{-1}\mathbf{\tilde x}$, with 
\begin{equation}
    D(\mathbf{x}, I_d) = \frac{Z_nZ_f}{Z_f - I_d(\mathbf{x})(Z_f - Z_n)},\quad \forall \mathbf{x}\in \Omega_f,
\end{equation}
where $\mathbf{\tilde x} = [x,y,1]^\top$ is the homogeneous representation of an image point $\mathbf{x} = [x,y]^\top$.

In~\cite{prisacariu12} it has been shown that it is beneficial not only to consider the points on the surface closest to the camera but also the most distant ones (on the backside of the object) for pose optimization. In order to obtain the respective coordinates for each pixel, we compute an additional reverse depth map $I_d^{r}$, for which we simply invert the OpenGL depth check used to compute the corresponding $Z$-buffer (see Figure~\ref{fig:depthmaps}). Given $I_d^{r}$, the farthest surface point $\mathbf{X}'(\mathbf{x}, I_d^{r})$ corresponding to a pixel $\mathbf{x}$ is also recovered as $\mathbf{X}'(\mathbf{x}, I_d^{r}) = D(\mathbf{x}, I_d^{r})K^{-1}\mathbf{\tilde x}$.

\subsection{Optimization} \label{sec:implopti}

We perform pose optimization hierarchically within a three level image pyramid generated with a down-scale factor of 2. The third level thereby corresponds to the camera matrix $\nicefrac{1}{4}K$, the second to $\nicefrac{1}{2}K$, and the first to $K$. In our current real-time implementation we first perform four iterations on the third, followed by two iterations on the second and finally a single iteration on the first level, i.e. the original full image resolution. In case of multiple objects all poses are updated sequentially once per iteration.

Each iteration starts by rendering and downloading the common silhouette mask $I_s$ and depth map $I_d$ as well as individual reverse depth maps $(I_d^{r})^j$ based on the current pose estimates $T^j$. To distinguish multiple objects, we render each model silhouette region $\Omega_f^j$ using a unique intensity corresponding to the model index $j$. Here, hierarchical rendering is achieved by simply adjusting the width and height of the OpenGL viewport according to the current pyramid level. Next, the individual signed distance transforms $\Phi^j$ are computed from $I_s$. For this we have implemented the efficient two-pass algorithm of~\cite{felzenszwalb12} in parallel on the CPU. Here, the first pass runs in parallel for each row and the second pass for each column of pixels. In addition to the distance value, we also store the 2D coordinates of the closest contour point $\mathbf{c}^j \in \mathbf{C}^j$ to every pixel $\mathbf{x} \in \Omega_b^j$. This is required for obtaining the corresponding 3D surface point needed to calculate the derivatives of $\Phi^j(\mathbf{x})$~\eqref{eq:jacobianvector} with respect to a background pixel. 

Finally, the Hessian and the gradient of the energy needed for the parameter step~\eqref{eq:updatestep} are calculated in parallel for each row of pixels on the CPU. Here, each thread calculates its own sums over $\psi(\mathbf{x})^{j}(J^j)^\top J^j$  and $J^j$ which are finally added up in the main thread. Following PWP3D, for each pixel we add both the Jacobian terms with respect to the coordinates of $\mathbf{X}'(\mathbf{x}, I_d)^{j}$ as well as $\mathbf{X}'(\mathbf{x}, (I_d^{r})^{j})^{j}$. For a further speed-up, we exploit that the Hessian is symmetrical, meaning that we only have to calculate the upper triangular part of it. The update step $\Delta \xi^j$ is then computed using Cholesky decomposition.

In our current implementation we choose $s = 1.2$ within Heaviside function $H_e$~\eqref{eq:heaviside} regardless of the pyramid level. We therefore always only need to perform pose optimization, i.e. compute the derivatives of each cost function $E^j$ within a band of $\pm8$\,px around each contour $\mathbf{C}^j$, i.e. $\forall \mathbf{x} \in \Omega$ with $\Phi^j(\mathbf{x}) \in [-8,8]$ (see Figure~\ref{fig:occlusion}). For other distances the corresponding Dirac delta value $\delta_e$ becomes very small. Since $\delta_e$ scales all other derivatives per pixel (see~\eqref{eq:j}), those outside this narrow band have a neglectable influence on the overall optimization. This further allows to restrict the processed pixels to a 2D ROI (region of interest) containing this contour band for an additional speed-up. We obtain this ROI by computing the 2D bounding rectangle of the projected bounding box of a model expanded by 8 pixels in each direction. This is done efficiently on the CPU without performing a full rendering. 

Due to the multi-scale strategy, it can easily happen that an object region only projects to a small amount of pixels in higher pyramid level at far distances to the camera. The derivatives computed from such few pixels can typically be less trusted and thus often move the optimization in the wrong direction. To encounter this effect we compute the area of the 2D bounding and check if it is too small at the current pyramid level (we use 3000 pixels as lower bound for an image resolution of $640\times512$\,px) at the beginning of each iteration. If this is the case, we directly move to the next higher image resolution in the pyramid and compute the pose update there.

\subsection{TCLC-Histograms}

We use the RGB color model with a quantization of 32 bins per channel to represent the tclc-histograms. The key idea to efficiently build and update the localized appearance model is to process each histogram region $\Omega_i^j$ in parallel on the CPU using Bresenham circles to scan the corresponding pixels. When updating the tclc-histograms we use learning rates of $\alpha_f = 0.1$ and $\alpha_b = 0.2$, allowing fast adaptation to dynamic changes. Based on the results presented in~\cite{hexner16}, we choose the histogram region radius as $r = 40$\,px for an image resolution of $640\times512$\,px, regardless of the object's distance to the camera. 

The reason why we are using a fixed radius is that for continuous tracking, we can only compute the segmentation within the histogram regions belonging to the silhouette in the previous frame. Thus, in cases of fast translational movement of the object or rotation of the camera it is possible that the object in the current frame projects entirely outside the previous histogram regions. This becomes more likely for smaller histogram radii. Therefore, as the radius shrinks with distance to the camera, the object is more likely to get lost at far distances. Non overlapping histogram regions in case of close distances and sparse surface sampling hardly influence the reliability of our approach. 

The other extreme case is when the object is so far away that all histograms overlap. Here, the discriminability of the appearance model is reduced since all pixels then lie within all histograms. The segmentation model then acts like the global approach constrained to a local region around the silhouette extended by the histogram radius. However this is still better than the global model and in our experience to be preferred over the increased risk of loosing the object easily with smaller radii.

After each pose optimization we compute the new 2D histogram centers by projecting all mesh vertices $\mathbf{X}_i^j$ of each model onto pixels $\mathbf{x}_i^j$. In practice we consider those with $\mathbf{x}_i^j \in \mathbf{C}^j$ as well as $d(\mathbf{x}_i^j, \mathbf{C}^j) \leq \lambda r$ (we use $\lambda = 0.1$), in order to ensure that the contour is evenly sampled. For runtime reasons, since this can lead to a large number of histograms that have to be updated, we randomly pick a maximum of 100 centers per frame. This Monte Carlo approach requires the mesh vertices $\mathbf{X}_i^j$ to be uniformly distributed across the 3D model in order to evenly cover all regions. They should also be limited in number to ensure that all histograms will get updated regularly. We therefore use two different 3D mesh representations for the model. The original mesh is used to render exact silhouette views regardless of the mesh structure while a reduced (we use a maximum of 5000 vertices) and evenly sampled version of the model is used for computing the 2D centers of the histograms.

\subsection{Occlusion handling} \label{sec:occlusions}

\begin{figure}[!tp]
\centering
\begin{picture}(127,100)
\put(0,0){\includegraphics[width=0.496\columnwidth]{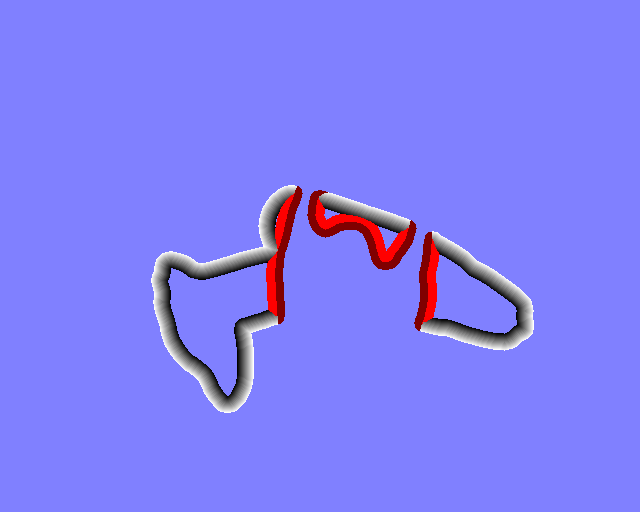}}
\footnotesize

\put(4,90){\textcolor{white}{$\Phi^1(\mathbf{x})$}}

\end{picture}\hfill
\begin{picture}(125,100)
\put(0,0){\includegraphics[width=0.496\columnwidth]{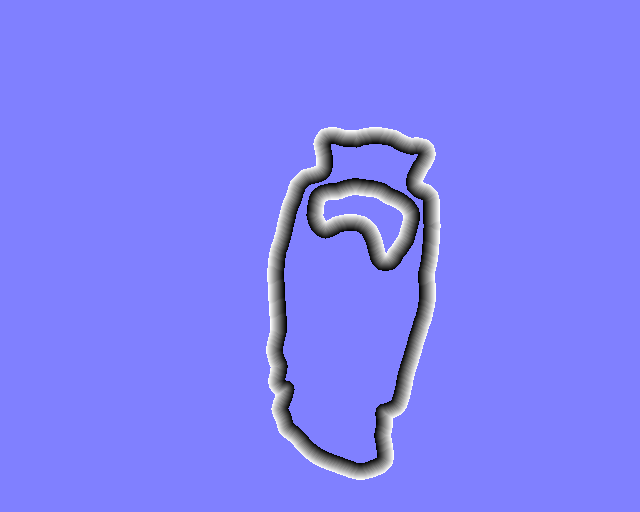}}
\footnotesize

\put(4,90){\textcolor{white}{$\Phi^2(\mathbf{x})$}}

\end{picture}
\caption{The two level-sets corresponding to $\mathbf{C}^1$ and $\mathbf{C}^2$ of Figure~\ref{fig:multimask}, visualized in the $\pm$8\,px band around the contours (grey pixels). Here, the distance values of $\Phi^1(\mathbf{x})$ that are influenced by the occlusion of $\mathbf{C}^1$ are marked red (bright inside and dark outside of $\Omega_f^1$).}
\label{fig:occlusion}
\end{figure}

As previously shown in Figure~\ref{fig:multimask}, when tracking multiple objects simultaneously, mutual occlusions are very likely to emerge. These must be handled appropriately on a per pixel level for pose optimization. In our approach occlusions can be detected with help of the common silhouette mask $I_s$ due to the $Z$-buffer of OpenGL. Thus, the respective contours $C^j$ computed directly from $I_s$, can contain segments resulting from occlusions that are considered in the respective signed distance transform. To handle this, for each object all pixels with a distance value that was influenced by occlusion have to be discarded for pose optimization (see Figure~\ref{fig:occlusion}). 

A straightforward approach as realized in~\cite{prisacariu12} is to render each model's silhouette $I_s^j$ separately as well as the common silhouette mask $I_s$.  The signed distance transforms $\Phi^j$ are then computed from the non-occluded $I_s^j$ where $I_s$ is only used to identify whether a pixel belongs to a foreign object region and thus has to be discarded.

Although this strategy is easy to compute it does not scale well with the number of objects since $I_s^j, I_d^j$, $(I_d^{r})^j, j=1,\ldots , m$ and $I_s$ have to be rendered and transferred to host memory in each iteration. In order to minimize rendering and memory transfer we follow the approach of~\cite{tjaden16}. Thus, we instead render the entire scene once per iteration and download the common silhouette mask $I_s$ and the respective depth-buffer $I_d$. The individual level-sets $\Phi^j$ are then directly computed from $I_s$. In addition to this we only have to render each model once separately in order to obtain the individual reverse depth buffers $(I_d^{r})^j$. This is not possible in a common scene rendering because the reverse depths of the occluded object would overwrite those of the object in front.

By only using $I_s$ and $I_d$ the detection of pixels with a distance value that was influenced by occlusion is split into two cases. For a pixel $\mathbf{x} \in \Omega_b^j$ outside of the silhouette region, i.e. $\Phi^j(\mathbf{x}) > 0$, we start by checking whether $I_s(\mathbf{x})$ equals another object index.
If so, $\mathbf{x}$ is discarded if also the depth at $I_d(\mathbf{x})$ of the other object is smaller than that of the closest contour pixel to $\mathbf{x}$, meaning that the other surface is actually in front of the current object (indicated with dark red in Figure~\ref{fig:occlusion}). For $\mathbf{x} \in \Omega_f^j$ inside of the silhouette region, i.e. $\Phi^j(\mathbf{x}) \leq 0$ we perform the same checks for all neighboring pixels outside of $\Omega_f^j$ to the closest contour pixel to $\mathbf{x}$. If any of these pixels next to the contour passes the mask and depth checks, $\mathbf{x}$ is discarded (indicated with bright red in Figure~\ref{fig:occlusion}).

\section{Experimental Evaluation} \label{sec:eval}

\begin{figure*}[tp]
\centering
\begin{picture}(335,130)

\put(0,0){\includegraphics[width=0.6\textwidth]{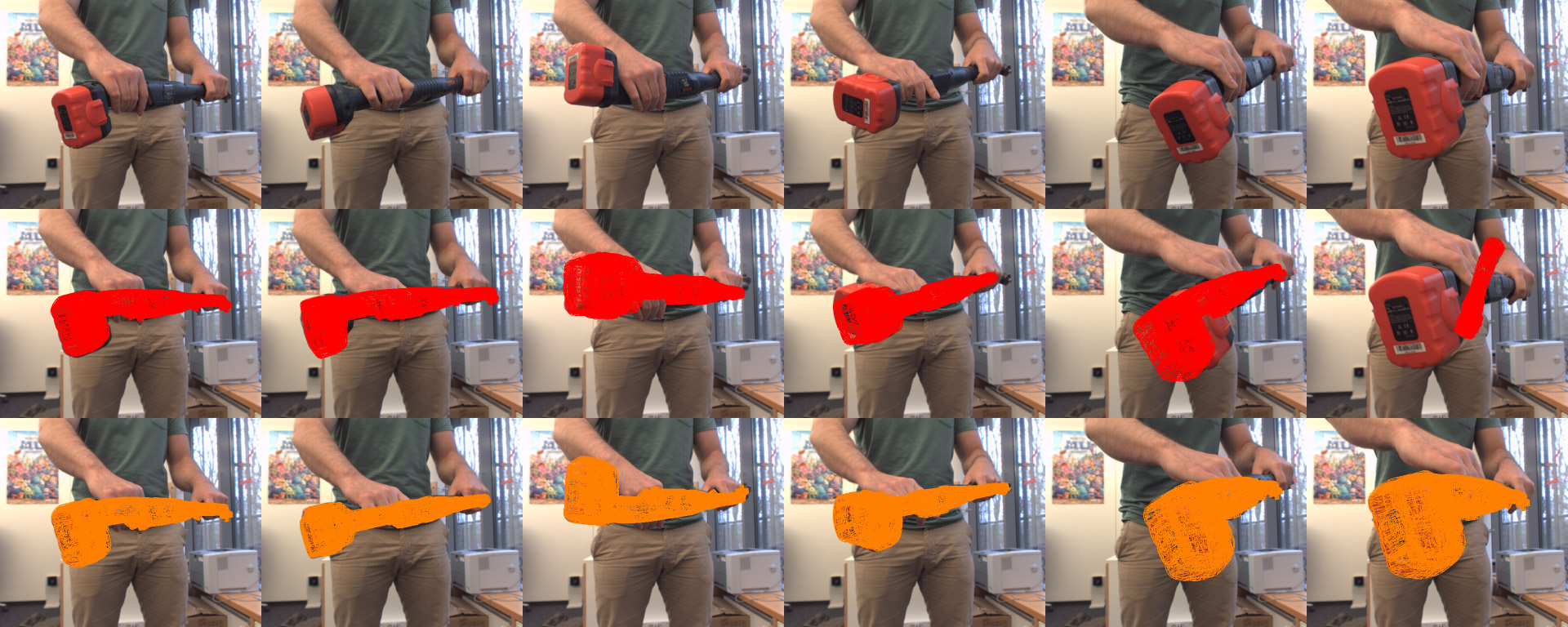}}

\begin{footnotesize}

\put(1, 85){\textcolor{white}{150}}
\put(53.5, 85){\textcolor{white}{230}}
\put(105.5, 85){\textcolor{white}{300}}
\put(157.5, 85){\textcolor{white}{350}}
\put(209, 85){\textcolor{white}{528}}
\put(261, 85){\textcolor{white}{586}}

\end{footnotesize}

\end{picture}
\hfill
\begin{footnotesize}
\input{pwp3d}
\hfill
\end{footnotesize}

\caption{Left: Visual results of a pose tracking experiment, where a hand-held screwdriver was rotated $360^\circ$ around its $X$-axis (top row: examples from the input sequence with the frame numbers depicted, middle row: results of the original first-order gradient descent implementation of PWP3D~\cite{prisacariu12}, bottom row: results of the Gauss-Newton-like optimization of~\cite{tjaden16}). Right: A corresponding plot of the estimated rotation around the $X$-axis for both methods.  While the oscillations around frames 105--150 and 500--600 indicate that the gradient descent time steps are already near the limit of stability, the algorithm cannot reliably estimate the full $360^\circ$ rotation leading to a failure of PWP3D (see last frame central row).
}
\label{fig:pwp3d}
\end{figure*}

In the following we present both quantitative and qualitative results of our approach in several different experiments. We start with an exemplary comparison between first-order and second-order optimization. This is followed by a comprehensive evaluation of tracking quality in the OPT and our novel dataset as well as complex mixed reality application examples. 

For all of these experiments we evaluate our implementation on a laptop with an Intel Core i7 quad core CPU @ 2.8 GHz and an AMD Radeon R9 M370X GPU. For the image shown in Figure~\ref{fig:segmentation} for example, a single optimization iteration at full 640$\times$512\,px resolution  takes $\sim$3.12\,ms (rendering: $\sim$23\%, level-set transform: $\sim$47\%, computing the Jacobians: $\sim$30\%) and the update of the tclc-histograms takes $\sim$3.8\,ms. However, the runtime of our method does not only depend on the number of objects but also on the size of the object region (i.e. distance to the camera), polygon count of the model and the number of histograms to be updated. In order to give an impression of the performance of our approach nonetheless, in the following we present the average runtimes for all sequences with different models and image resolutions. 

\subsection{Comparison to PWP3D} \label{sec:pwp3d}

In~\cite{tjaden16} the advantages of the Gauss-Newton-like second-order strategy in comparison to the first-order gradient descent method of the original PWP3D~\cite{prisacariu12} were demonstrated. Essentially, the here proposed true Gauss-Newton optimization has similar convergence properties to the previous Gauss-Newton-like one. We therefore include an exemplary experiment from~\cite{tjaden16} in order to illustrate the general difference between first-order and second-order optimization in this context (see Figure~\ref{fig:pwp3d}). Note that both approaches use the same appearance model based on global color histograms. 

In the selected experiment a cordless screwdriver was tracked while being moved in front of a stationary camera. The results show the dependence of the step-sizes in PWP3D on the distance to the camera. Here it holds, if the distance between object and camera becomes too small, the step-sizes are too large and the pose starts to oscillate. However, if this distance increases the overall optimization quality degrades, resulting in the step sizes to be too small to converge. 

The sequence contains a challenging full $360^\circ$ turn around the $X$-axis of the screwdriver (e.g. frames 180--400). For this the rotation step-size for PWP3D was set to a large value such that it was close to oscillating (e.g. frames 105--150) since this produced the best overall results.
While the Gauss-Newton-like strategy is able to correctly track the entire motion, PWP3D fails to determine the rotation of the object despite the large step-size for rotation.
Starting at around frame 450, the screwdriver was moved closer towards the camera, leading to a tracking loss of PWP3D at frame 586, while the Gauss-Newton-like method remained stable. 

Due to these essential drawbacks we did not include PWP3D in the quantitative evaluation within our new complex dataset in Section~\ref{sec:results}. It would require to manually set three different step-sizes individually for each object and we did not see a chance of it performing competitively.

\subsection{The RBOT Dataset}\label{sec:ourdataset}

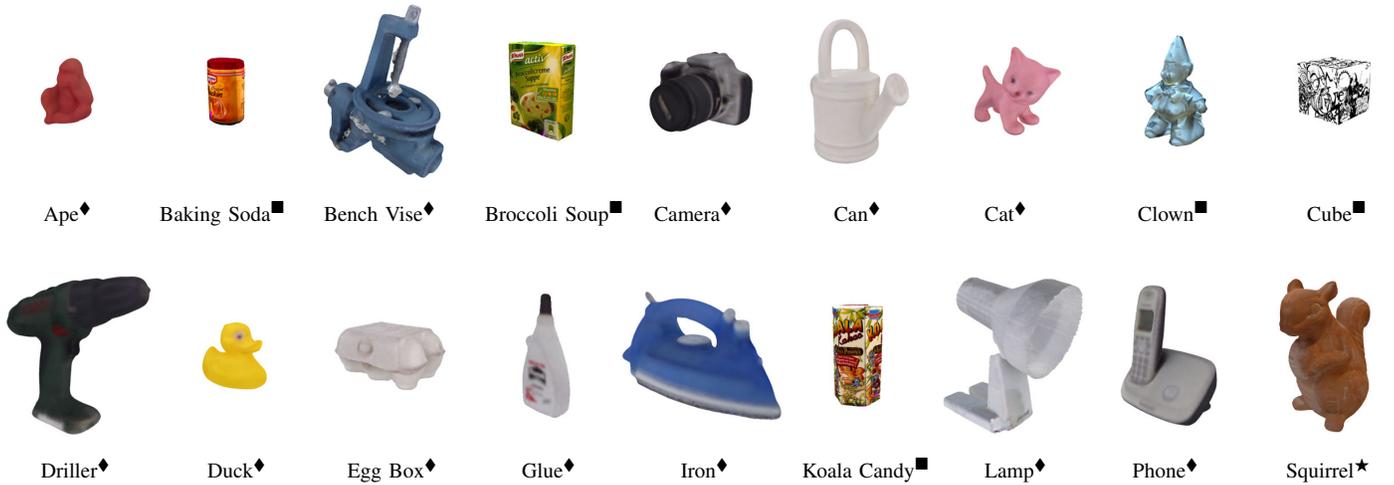
\begin{figure*}[!htp] 
    \centering
    
    \begin{picture}(514,178)

    \put(0,0){\includegraphics[width=\textwidth]{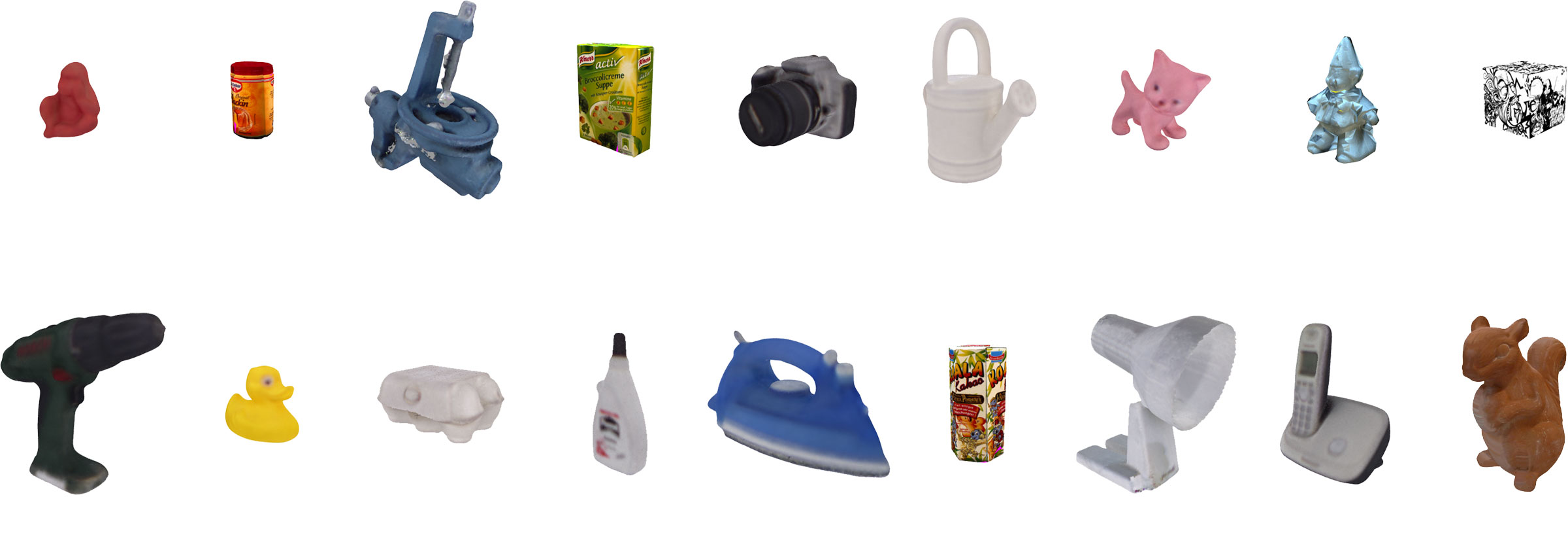}\hfill}
    
    \begin{footnotesize}
    
    \put(14, 97){\textcolor{black}{Ape$^\blacklozenge$}}
    \put(58, 97){\textcolor{black}{Baking Soda$^\blacksquare$}}
    \put(120, 97){\textcolor{black}{Bench Vise$^\blacklozenge$}}
    \put(181, 97){\textcolor{black}{Broccoli Soup$^\blacksquare$}}
    \put(245, 97){\textcolor{black}{Camera$^\blacklozenge$}}
    \put(313, 97){\textcolor{black}{Can$^\blacklozenge$}}
    \put(370, 97){\textcolor{black}{Cat$^\blacklozenge$}}
    \put(428, 97){\textcolor{black}{Clown$^\blacksquare$}}
    \put(492, 97){\textcolor{black}{Cube$^\blacksquare$}}
    
    \put(13, 0){\textcolor{black}{Driller$^\blacklozenge$}}
    \put(76, 0){\textcolor{black}{Duck$^\blacklozenge$}}
    \put(129, 0){\textcolor{black}{Egg Box$^\blacklozenge$}}
    \put(195, 0){\textcolor{black}{Glue$^\blacklozenge$}}
    \put(255, 0){\textcolor{black}{Iron$^\blacklozenge$}}
    \put(301, 0){\textcolor{black}{Koala Candy$^\blacksquare$}}
    \put(370, 0){\textcolor{black}{Lamp$^\blacklozenge$}}
    \put(426, 0){\textcolor{black}{Phone$^\blacklozenge$}}
    \put(484, 0){\textcolor{black}{Squirrel$^\bigstar$}}

    \end{footnotesize}
    
    \end{picture}
    \caption{An overview of all eighteen models included in the RBOT dataset. The well-textured models from the Rigid Pose tracking dataset~\cite{pauwels13} are marked with$^\blacksquare$. The weakly-textured models from the LINE-MOD detection dataset~\cite{hinterstoisser12a} are marked with$^\blacklozenge$. Here, all models are rendered at the same pose for scale comparison.} 
    \label{fig:allmodels} 
\end{figure*}

We call the proposed semi-synthetic monocular 6DOF pose tracking dataset \textit{RBOT} (\textit{Region-based Object Tracking}) in regard to the proposed method. We have made it publicly available for download under: \url{http://cvmr.info/research/RBOT}. It comprises a total number of eighteen different objects, all available as textured 3D triangle meshes. In addition to our own model of a squirrel clay figurine, we have included a selection of twelve models from the LINE-MOD dataset~\cite{hinterstoisser12a} and five from the Rigid Pose dataset~\cite{pauwels13} as shown in Figure~\ref{fig:allmodels}. 

\begin{figure}[!tp]
\centering
\begin{picture}(127,100)
\put(0,0){\includegraphics[width=0.496\columnwidth]{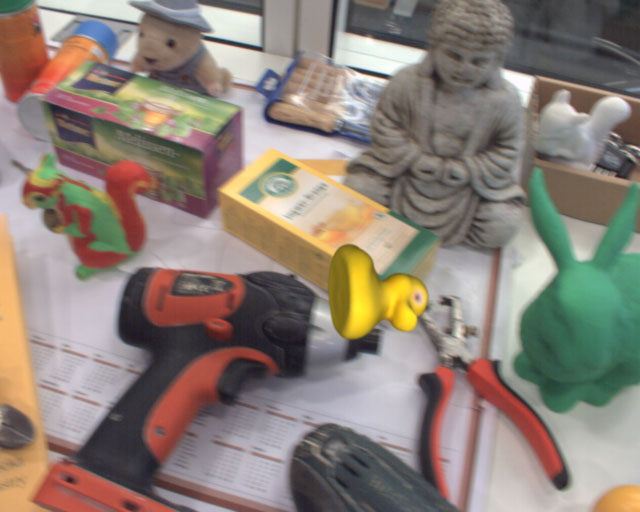}}
\footnotesize

\put(95,91){\textcolor{white}{Regular}}

\end{picture}\hfill
\begin{picture}(125,100)
\put(0,0){\includegraphics[width=0.496\columnwidth]{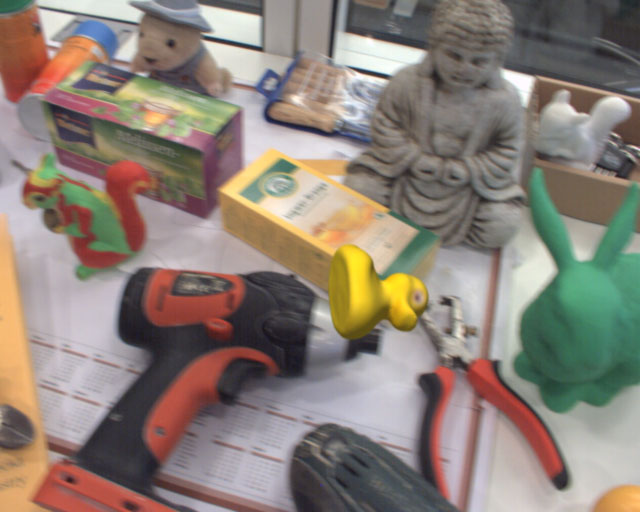}}
\footnotesize

\put(74,91){\textcolor{white}{Dynamic light}}

\end{picture}\hfill\\[2pt]
\begin{picture}(127,100)
\put(0,0){\includegraphics[width=0.496\columnwidth]{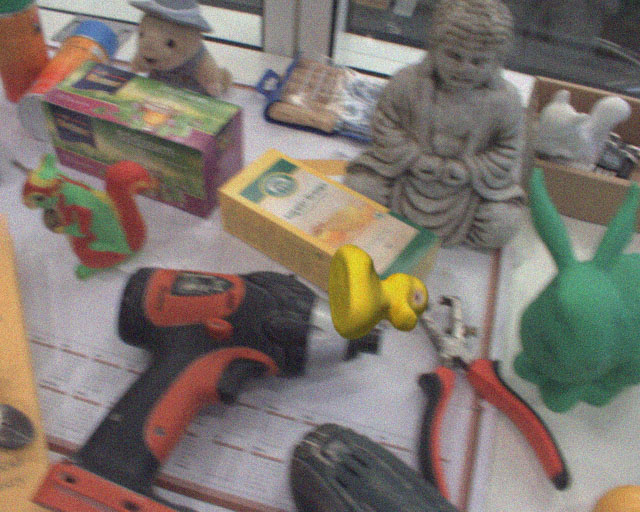}}
\footnotesize

\put(102,91){\textcolor{white}{Noisy}}

\end{picture}\hfill
\begin{picture}(125,100)
\put(0,0){\includegraphics[width=0.496\columnwidth]{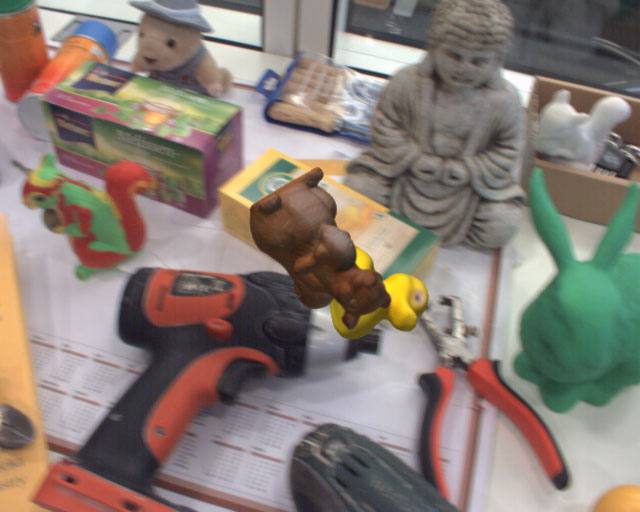}}
\footnotesize

\put(89,91){\textcolor{white}{Occlusion}}

\end{picture}
\caption{An example frame from the RBOT dataset with the duck model in the four different variants. Visually comparing the regular with the dynamic light version clarifies the impact of the lighting aspect on the appearance of the object region.}
\label{fig:rbot}
\end{figure}
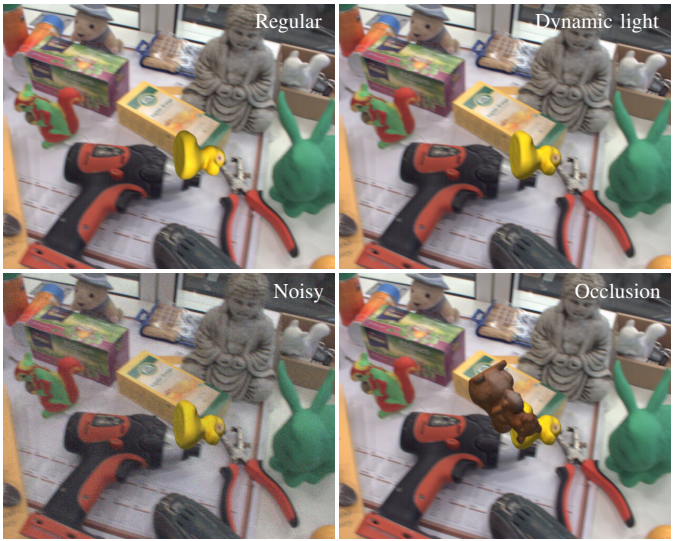

For each model we have generated four variants of a semi-synthetic image sequence with increasing complexity (see Figure~\ref{fig:rbot}). The first \textit{regular} variant contains a single object rendered with respect to a static point light source located above the origin of the virtual camera. This simulates a moving object in front of a static camera. The second variant is the same as the first but with a dynamic light source in order to simulate simultaneous motion of both the object and the camera. The images in the third variant were also rendered with respect to the moving light source and further distorted by adding artificial Gaussian noise. Finally, the forth variant contains an additional second object (the squirrel) which orbits around the first object and thus frequently occludes it. These multi-object sequences also include the dynamic light source.

Regardless of the object and the variant, we animated the model using the same pre-defined trajectory of continuous 6DOF motion in all sequences. We also always use the same background video for compositing. This video was captured by moving a hand-held camera arbitrarily in a cluttered desktop scene. In order to increase realism, we rendered the models with anti-aliasing and blurred the object regions in the composition using a $3\times3$ Gaussian kernel. The latter in particular smooths the transition between the object and the background, blending it more realistically with the rest of the scene. Each sequence contains 1001 RGB frames of $640\times512$\,px resolution, where the first frame is always used for initialization. This results in a total number of $72000 = 18\cdot4\cdot1000$ color images. For each frame we provide the ground truth poses for the two objects as well as the intrinsic camera matrix used for rendering which we obtained from calibrating the camera that recorded the background video.

\subsection{Experimental Results} \label{sec:results}

In the following we present tracking results for our approach within the mentioned OPT and the proposed RBOT dataset.

\textbf{OPT Dataset} -- We evaluated our method within the OPT dataset~\cite{wu17} by using all RGB image sequences provided for each of the six 3D objects (Bike, Chest, House, Ironman, Jet and Soda) at $1920\times1080$\,px resolution. To compare our results to those provided in~\cite{wu17}, we conducted the experiment in the same fashion and measure the pose error as
\begin{equation}
    e_k = \frac{1}{n}\sum_{i = 1}^{n}\|\left(T(t_k)\mathbf{\tilde X}_i - T_{gt}(t_k)\mathbf{\tilde X}_i\right)_{3 \times 1}\|_2,
    \label{eq:opterror}
\end{equation}
for each frame, with $T_{gt}(t_k)$ being the corresponding ground truth pose. Pose tracking is considered successful when $e_{k} < \lambda_{e}d_{m}$, where $d_{m}$ is the diameter (i.e. the largest distance between vertices) of the model and $\lambda_{e}$ is a predefined threshold. The overall tracking quality of our method is then computed in form of an AUC (area under curve) score, where the percentage amount of successfully tracked poses across all frames is integrated for all $\lambda_{e} \in [0,0.2]$ per model. For this experiment ground truth pose information was only used at the start of each of the different sequences to initialize the tracking. Here, tracking losses were ignored, meaning that the pose was never reset to ground truth during a sequence.

\begin{table}[htp]
\setlength{\tabcolsep}{4.15pt}
\renewcommand{\arraystretch}{1.1}
\begin{tabular}{ | c | c  c  c  c  c  c  |}
\hline
\textbf{Method} & \rotatebox{90}{Bike} & \rotatebox{90}{Chest} & \rotatebox{90}{House} & \rotatebox{90}{Ironman} & \rotatebox{90}{Jet} & \rotatebox{90}{Soda}\\
\hline
PWP3D~\cite{prisacariu12} & 5.358 & 5.551 & 3.575 & 3.915 & 5.813 & 5.870\\
UDP~\cite{brachmann16} & 6.097 & 6.791 & 5.974 & 5.250 & 2.342 & 8.494\\
ElasticFusion~\cite{whelan15} & 1.567 & 1.534 & 2.695 & 1.692 & 1.858 & 1.895\\
ORB-SLAM2~\cite{murartal17} & 10.410 & \textbf{15.531} & \textbf{17.283} & 11.198 & 9.931 & \textbf{13.444}\\
Proposed & \textbf{11.903} & 11.764 & 10.150 & \textbf{11.986} &  \textbf{13.217} &  8.861\\
\hline
Avg. Runtime & 62.4 & 62.6 & 62.4 & 62.4 & 63.4 & 65.9  \\
\hline
\end{tabular}
\caption{AUC scores (higher is better) of our approach in the OPT dataset compared to the results presented in~\cite{wu17}. In the last row also the average runtime of our approach (in ms) per object is given. The experiment confirms that the proposed solution performs significantly better than PWP3D, UDP and ElasticFusion and even outperforms ORB-SLAM2 is many cases.}
\label{tab:optresults}
\end{table}

The resulting AUC scores are shown in Table~\ref{tab:optresults} in comparison to those available in~\cite{wu17} for PWP3D~\cite{prisacariu12}, UDP~\cite{brachmann16}, ElasticFusion~\cite{whelan15} and ORB-SLAM2~\cite{murartal17}. However, apart from PWP3D none of the other three methods are dedicated object pose tracking solutions. UDP is a monocular RGB object pose detection method while ElasticFusion and ORB-SLAM2 are visual SLAM approaches for camera pose localization in indoor and outdoor scenes which can, however, be applied in this static object scenario. ElasticFusion relies on RGB-D data, ORB-SLAM2 uses gradient-based corner features and both were restricted to the object region for the experiment. For more detailed information on the setup, results and runtimes of the other methods, please refer to~\cite{wu17}. 

The results show that the proposed method again performs significantly better than the original PWP3D algorithm as well as UDP and ElasticFusion. For the three least textured objects (Bike, Ironman and Jet), our approach even outperforms ORB-SLAM2, being among the state-of-the-art monocular SLAM algorithms. For the other three relatively well textured objects ORB-SLAM2 performs better because it uses features within the entire object region for pose estimation instead of being constrained to the silhouette. Here, the large image resolution generally helps to detect an increased number features even at larger distances which such approaches benefit from.

\begin{table*}[tp]
\setlength{\tabcolsep}{4.55pt}
\renewcommand{\arraystretch}{1.1}
\begin{tabular}{|c | c | c  c  c  c  c  c  c  c  c  c  c  c  c  c  c  c  c  c |}
\hline
\textbf{Variant} & \textbf{Method} & \rotatebox{90}{Ape} & \rotatebox{90}{Baking Soda} & \rotatebox{90}{Bench Vise} & \rotatebox{90}{Broccoli Soup} & \rotatebox{90}{Camera} & \rotatebox{90}{Can} & \rotatebox{90}{Cat} & \rotatebox{90}{Clown} & \rotatebox{90}{Cube} & \rotatebox{90}{Driller} & \rotatebox{90}{Duck} & \rotatebox{90}{Egg Box} & \rotatebox{90}{Glue} & \rotatebox{90}{Iron} & \rotatebox{90}{Koala Candy} & \rotatebox{90}{Lamp} & \rotatebox{90}{Phone} & \rotatebox{90}{Squirrel}\\
\hline
\multirow{2}{*}{Regular} & \cite{tjaden17} & 62.1 & 30.5 & 95.8 & 66.2 & 61.6 & 81.7 & \textbf{96.7} & 89.1 & 44.1 & 87.7 & 74.9 & 50.9 & 20.2 & 68.4 & 20.0 & 92.3 & 64.9 & 98.5  \\
& Proposed & \textbf{85.0} & \textbf{39.0} & \textbf{98.9} & \textbf{82.4} & \textbf{79.7} & \textbf{87.6} & 95.9 & \textbf{93.3} & \textbf{78.1} & \textbf{93.0} & \textbf{86.8} & \textbf{74.6} & \textbf{38.9} & \textbf{81.0} & \textbf{46.8} & \textbf{97.5} & \textbf{80.7} & \textbf{99.4}  \\
\hline
\multirow{2}{*}{Dynamic Light} & \cite{tjaden17} & 61.7 & 32.0 & 94.2 & 66.3 & 68.0 & 84.1 &\textbf{96.6} & 85.8 & 45.7 & 88.7 & 74.1 & 56.9 & 29.9 & 49.1 & 20.7 & 91.5 & 63.0 & 98.5  \\ 
& Proposed & \textbf{84.9} & \textbf{42.0} & \textbf{99.0} & \textbf{81.3} & \textbf{84.3} & \textbf{88.9} & 95.6 & \textbf{92.5} & \textbf{77.5} & \textbf{94.6} & \textbf{86.4} & \textbf{77.3} & \textbf{52.9} & \textbf{77.9} & \textbf{47.9} & \textbf{96.9} & \textbf{81.7} & \textbf{99.3}  \\
\hline
Noisy +& \cite{tjaden17} & 55.9 & 35.3 & 75.4 & 67.4 & 27.8 & 10.2 & 94.3 & 33.4 & 8.6 & 50.9 & 76.3 & 2.3 & 2.2 & 18.2 & 11.4 & 36.6 & 31.3 & 93.5  \\
Dynamic Light & Proposed & \textbf{77.5} & \textbf{44.5} & \textbf{91.5} & \textbf{82.9} & \textbf{51.7} & \textbf{38.4} & \textbf{95.1} & \textbf{69.2} & \textbf{24.4} & \textbf{64.3} & \textbf{88.5} & \textbf{11.2} & \textbf{2.9} & \textbf{46.7} & \textbf{32.7} & \textbf{57.3} & \textbf{44.1} & \textbf{96.6}  \\
\hline
Unmodelled & \cite{tjaden17} & 55.2 & 29.9 & 82.4 & 56.9 & 55.7 & 72.2 & 87.9 & 75.7 & 39.6 & 78.7 & 68.1 & 47.1 & 26.2 & 35.6 & 16.6 & 78.6 & 50.3 & 77.6  \\
Occlusion & Proposed & \textbf{80.0} & \textbf{42.7} & \textbf{91.8} & \textbf{73.5} & \textbf{76.1} & \textbf{81.7} & \textbf{89.8} & \textbf{82.6} & \textbf{68.7} & \textbf{86.7} & \textbf{80.5} & \textbf{67.0} & \textbf{46.6} & \textbf{64.0} & \textbf{43.6} & \textbf{88.8} & \textbf{68.6} & \textbf{86.2}  \\
\hline
Modelled & \cite{tjaden17} & 60.3 & 31.0 & 94.3 & 64.5 & 67.0 & 81.6 & 92.5 & 81.4 & 43.2 & 89.3 & 72.7 & 51.6 & 28.8 & 53.5 & 19.1 & 89.3 & 62.2 & 96.7  \\
Occlusion & Proposed & \textbf{82.0} & \textbf{42.0} & \textbf{95.7} & \textbf{81.1} & \textbf{78.7} & \textbf{83.4} & \textbf{92.8} & \textbf{87.9} & \textbf{74.3} & \textbf{91.7} & \textbf{84.8} & \textbf{71.0} & \textbf{49.1} & \textbf{73.0} & \textbf{46.3} & \textbf{90.9} & \textbf{76.2} & \textbf{96.9}  \\
\hline
Avg. Runtime & Proposed & 15.5 & 15.7 & 20.3 & 16.6 & 17.4 & 18.9 & 16.9 & 15.9 & 16.8 & 19.4 & 15.7 & 16.8 & 16.1 & 19.1 & 16.5 & 21.8 & 17.6 & 19.1  \\
\hline
\end{tabular}
\caption{Tracking success rates (in \%) of the proposed method in comparison to that of~\cite{tjaden17} in the RBOT dataset. In the last row we furthermore denote the average runtime of our approach (in ms) measured across the first three image sequence variants.  These experiments confirm that the systematic derivation of a Gauss-Newton optimization suggested in Section \ref{sec:gn} leads to a significant performance improvement over \cite{tjaden17} for almost all objects.}
\label{tab:rbotresults}
\end{table*}

\textbf{RBOT Dataset} -- For evaluation within the RBOT dataset we denote the sequence of ground truth poses by $T^j_{gt}(t_k)$, composed of $R^j_{gt}(t_k)$ and $\mathbf{t}^j_{gt}(t_k)$ with $k = 0,\dots, 1000$. In contrast to the previous experiment, here we use a different, more generic criterion to measure the tracking error as suggested in~\cite{garon17}, since~\eqref{eq:opterror} highly depends on the object's geometry. Starting at $T^j(t_0) = T^j_{gt}(t_0)$, for each subsequent frame we thus compute the tracking error separately for translation $e_k^j(\mathbf{t}) = \|\mathbf{t}^j(t_k) - \mathbf{t}^j_{gt}(t_k)\|_2$ and rotation
\begin{equation}
    e_k^j(R) = \cos^{-1}\left(\frac{\trace(R^j(t_k)^\top R^j_{gt}(t_k))-1}{2}\right),
\end{equation}
for each object to evaluate the tracking success rate. If $e_k^j(\mathbf{t})$ is below 5\,cm and $e_k^j(R)$ below $5^\circ$, we consider the pose to be successfully tracked. Otherwise, if one of the errors is not within its boundaries, we consider the tracking to be lost and reset it to the ground truth pose, i.e. $T^j(t_k) = T^j_{gt}(t_k)$. In our experiments we evaluated the multi-object sequences in two different ways. We either track only the pose of the first (varying) object or both of them (the varying object and the occluding squirrel). When tracking only one of the objects, the occurring occlusions are \textit{unmodelled} as we call it here. These are much harder to handle than \textit{modelled} occlusions, where the pose and geometry of the occluding object is known, in case of tracking both objects.

In Table~\ref{tab:rbotresults} we present the tracking success rates of the proposed method and the method presented in~\cite{tjaden17} for all sequences in all variants. The results show that the novel re-weighted Gauss-Newton optimization outperforms the previous Gauss-Newton-like strategy in most cases by a large margin. Here, the biggest difference occurs in presence of noise. This can be explained by the proposed weighting strategy, since the $\psi(\mathbf{x})$ terms generally reduce the influence of false segmented (outlier) pixels. The experiments also demonstrate the robustness of the appearance model based on tclc-histograms towards a moving light source. For both methods, compared to the regular scenario, the performance often even improves in the dynamic light variant and hardly deteriorates otherwise. However, it can also be seen that both approaches perform significantly worse for objects with ambiguous silhouettes (e.g. Baking Soda,  Glue and Koala Candy) and struggle more with image noise in case of objects of a less distinct color (e.g. Camera, Can, Cube, Egg Box etc.).




\subsection{Applications to Mixed Reality}

\begin{figure}[!tp] \centering
    \includegraphics[width=0.497\columnwidth]{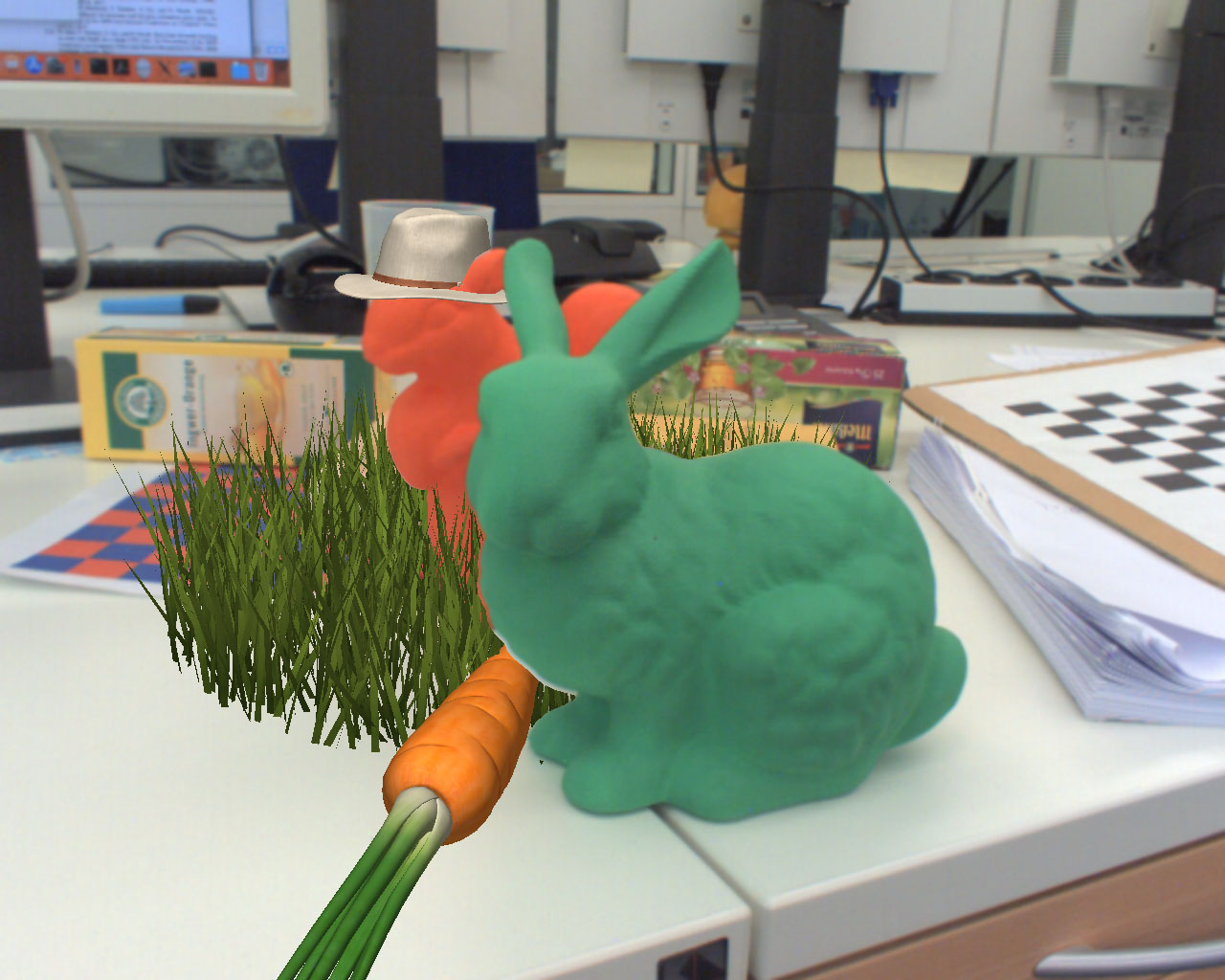}\hfill
    \includegraphics[width=0.497\columnwidth]{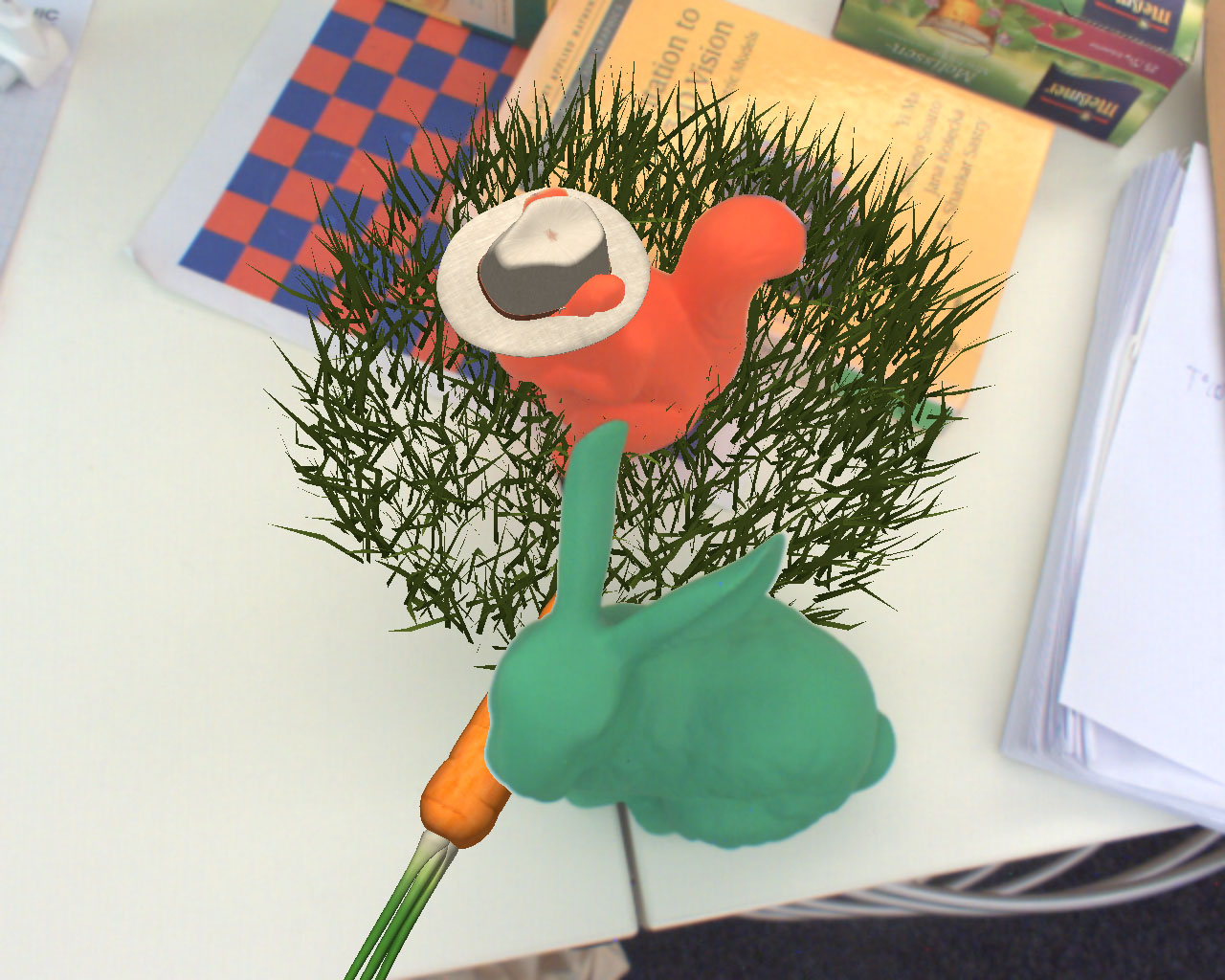}\hfill\\[0.4mm]
    \includegraphics[width=0.497\columnwidth]{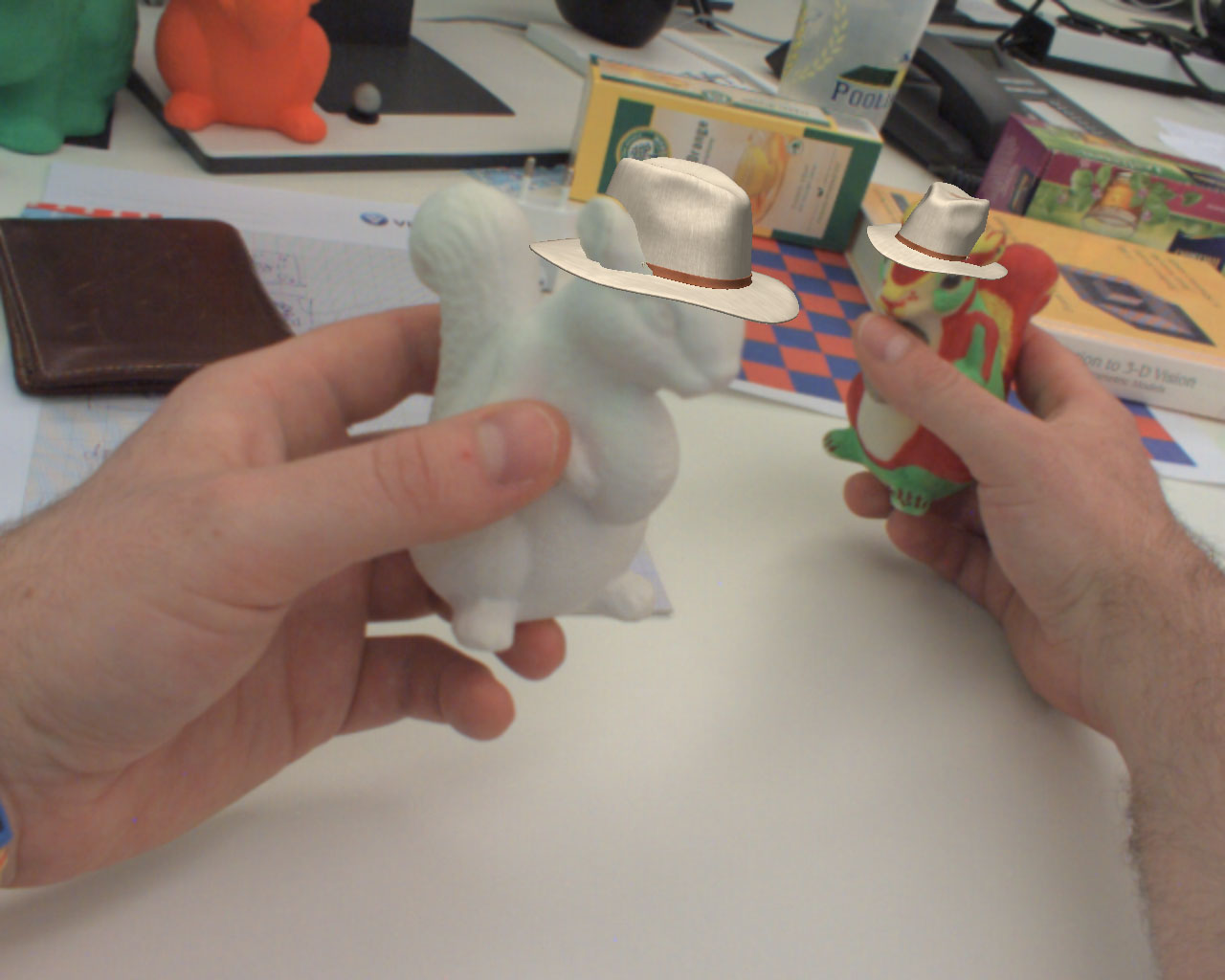}\hfill
    \includegraphics[width=0.497\columnwidth]{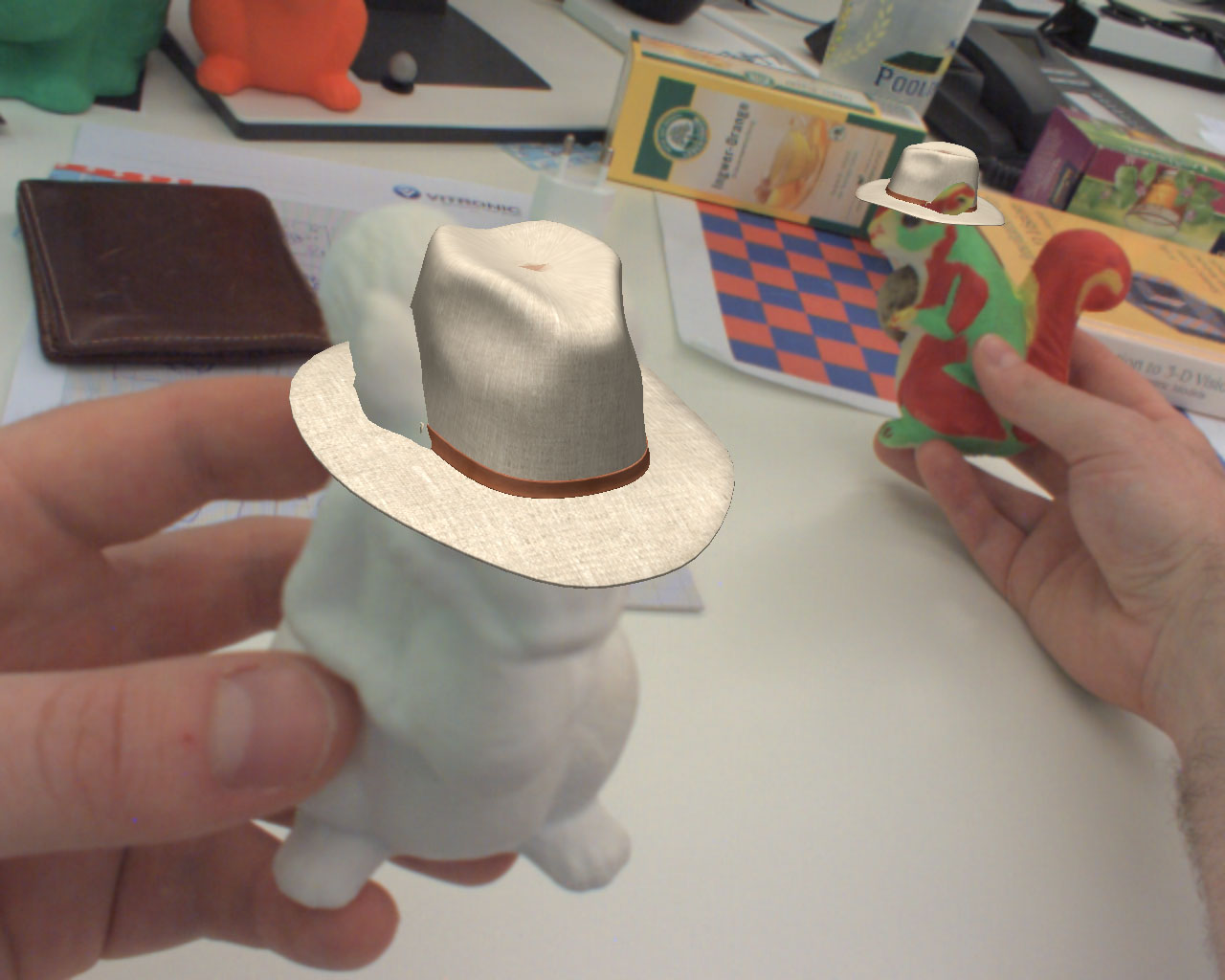}\hfill
    \caption{Mixed reality visualizations in two different complex scenes with two tracked objects each. Based on the pose estimates, the real objects are augmented with virtual attachments (hats, a carrot and a patch of grass) in real-time. Here, despite large perspective changes and both modelled and unmodelled occlusions, the augmentations remain precisely in place. This makes our method very attractive for complex mixed reality systems.} 
    \label{fig:ar} 
\end{figure}

Due to its low run-time and high accuracy our approach is very suitable for mixed reality applications. The ability to track multiple objects enables immersive visualizations in highly dynamic scenarios. Here, the virtual augmentations can realistically be occluded by the real objects since their geometry and poses are accurately known (see Figure~\ref{fig:ar}). What makes it even more attractive for mixed reality systems, is that our approach can handle fair amounts of unmodelled occlusion (e.g. by hands). Therefore, object-specific augmentations will remain in place while a user inspects objects by manipulating them manually. This further allows to turn arbitrary objects into 6DOF motion input devices.

\section{Conclusion} \label{sec:conc}

With this work we have closed two gaps in literature on 6DOF object pose tracking. Firstly, we have provided a fully analytic derivation of a Gauss-Newton optimization that was originally lacking in~\cite{tjaden16}.  It is derived in form of a re-weighted nonlinear least-squares estimation.  A systematic quantitative evaluation in Table \ref{tab:rbotresults} shows that the resulting update scheme leads to significant improvements over \cite{tjaden17}. Secondly, we have presented and created a novel large dataset for object tracking that covers practically relevant scenarios beyond prior comparable work. We believe that the community will benefit from both of these contributions.

However, regardless of the employed optimization strategy the presented approach only relies on the objects' contours to determine their poses. It is therefore prone to fail for objects with ambiguous silhouette projections such as bodies of revolution (e.g. Baking Soda and Koala Candy from the dataset). To cope with this restriction and further improve tracking accuracy a photometric term could be incorporated in the cost function with regard to the objects' texture. Assuming an object is sufficiently textured this should resolve the silhouette ambiguity in many cases.

Finally, we want to point out that the proposed dataset could be used to train or evaluate deep-learning approaches for 6DOF tracking in the future. This has become even more relevant recently, since our semi-synthetic images resemble those of~\cite{kehl17b, hinterstoisser17}.

\section{Acknowledgements} \label{sec:ack}
Part of this work was funded by the Federal Ministry for Economic Affairs and Energy (BMWi).
We thank Stefan Hinterstoisser and Karl Pauwels for letting us re-use their 3D models for our dataset.  DC was supported by the ERC Consolidator Grant 3DReloaded. 

{\small
\bibliographystyle{ieee}
\bibliography{references}
}

%

\begin{IEEEbiography}[{\includegraphics[width=2.5cm,clip,keepaspectratio]{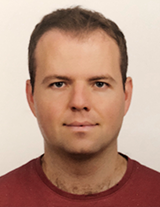}}]{Henning Tjaden is a PhD student at the Johannes Gutenberg University of Mainz and a research assistant at the RheinMain University of Applied Sciences in Germany where he is part of the Computer Vision and Mixed  Reality group. His current research interests include real-time object pose estimation of 3D objects from 2D images by using only a single monocular camera. He is especially interested
in improving the robustness, accuracy and runtime performance in order enable approaches to be applicable to a large variety of practical scenarios in dynamic human environments.} 
\end{IEEEbiography}

\begin{IEEEbiography}[{\includegraphics[width=2.5cm,clip,keepaspectratio]{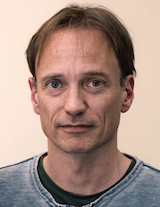}}]{Ulrich Schwanecke received a PhD in Mathematics (2000) from the Technische Universit\"at Darmstadt, Germany. He then spent a year as a postdoctoral researcher at Max Planck Institute for Informatics in Saarbr\"ucken, Germany. From 2001 until 2003 he was a permanent researcher at Daimler Chrysler Research in Ulm and Stuttgart, Germany. Since 2003 Prof. Schwanecke is a full professor for computer graphics and vision at RheinMain University of Applied Sciences in Wiesbaden, Germany. In 2017 he was awarded with the research award 
of the Hessian Universities of Applied Sciences.} 
\end{IEEEbiography}

\begin{IEEEbiography}[{\includegraphics[width=2.5cm,clip,keepaspectratio]{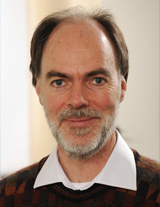}}]{Elmar Sch\"omer received a PhD (1994) and his venia legendi (1999) in Computer Science from Saarland University, Germany. He then spent two years as a senior researcher at the Max Planck Institute for  Informatics in Saarbr\"ucken. Since 2002 Prof. Sch\"omer is a full professor for computer graphics and computational geometry at the Johannes Gutenberg University of Mainz, Germany. His main areas of research include efficient (parallel) geometric algorithms for optimization and simulation 
problems.} 
\end{IEEEbiography}


\begin{IEEEbiography}[{\includegraphics[width=2.5cm,clip,keepaspectratio]{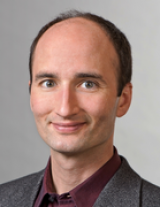}}]{Daniel
    Cremers received a PhD in Computer Science (2002)
    from the University of Mannheim, Germany.  He spent
    two years as a postdoctoral researcher at the UCLA and one year as
    a permanent researcher at Siemens Corporate Research in Princeton.
    From 2005 until 2009 Prof. Cremers headed the Computer Vision
    Group at the University of Bonn, Germany. Since 2009 he is full
    professor for informatics and mathematics at TU
    Munich. He received several awards, including the
    {\em Olympus Award 2004},  the {\em 2005 UCLA Chancellor's
      Award} and the Gottfried Wilhelm Leibniz Award 2016.}
\end{IEEEbiography} 




\end{document}